\documentclass[aps,prl,twocolumn,superscriptaddress,floatfix]{revtex4-2}
\usepackage{amsmath}
\usepackage{graphicx}
\usepackage{color}
\usepackage{comment}
\usepackage{hyperref}
\usepackage{subcaption}
\usepackage{float}  
\usepackage{natbib}
\usepackage{array}
\usepackage{tabularx}   
\usepackage{booktabs}   

\begin{document}

\title{Supervision policies can shape long-term risk management in\\ general-purpose AI models}

\author{Manuel Cebrian}
\affiliation{Department of Statistics, Universidad Carlos III de Madrid, Spain} 
\affiliation{Center for Automation and Robotics, Spanish National Research Council, Madrid, Spain}
\

\author{Emilia Gómez}
\affiliation{European Commission, Joint Research Centre, Seville, Spain}

\author{David {Fernández Llorca}}
\affiliation{European Commission, Joint Research Centre, Seville, Spain}
\affiliation{Computer Engineering Department, University of Alcalá, Alcalá de Henares, Spain}

\begin{abstract} 
The rapid proliferation and deployment of General-Purpose AI (GPAI) models, including large language models (LLMs), present unprecedented challenges for AI supervisory entities. We hypothesise that these entities will need to navigate an emergent ecosystem of risk and incident reporting, likely to exceed their supervision capacity. To investigate this, we develop a simulation framework parametrised by features extracted from the diverse landscape of risk, incident or hazard reporting ecosystems—including community-driven platforms, crowdsourcing initiatives, and expert assessments. We evaluate four supervision policies: non-prioritised (first-come, first-served), random selection, priority-based (addressing the highest-priority risks first), and diversity-prioritised (balancing high-priority risks with comprehensive coverage across risk types). Our results indicate that while priority-based and diversity-prioritised policies are more effective at mitigating high-impact risks—particularly those identified by experts—they may inadvertently neglect systemic issues reported by the broader community. This oversight can create feedback loops that amplify certain types of reporting while discouraging others, leading to a skewed perception of the overall risk landscape. We validate our simulation results with several real-world datasets, including one with over a million ChatGPT interactions, including more than 150K conversations identified as risky. This validation underscores the complex trade-offs inherent in AI risk supervision and highlights how the choice of risk management policies can shape the future landscape of AI risks across diverse GPAI models used in society.
\end{abstract}

\maketitle

\section{Introduction}

The rapid adoption of GPAI models—including but not limited to LLMs—across various applications has introduced significant opportunities and unprecedented challenges. These models, capable of generating human-like text and handling a wide range of tasks across diverse data types, are increasingly integrated into everyday systems, raising key concerns related to cybersecurity, bias, privacy violations, misinformation, and the misuse of AI to generate harmful content \cite{brown2020language,vaswani2017attention, vaswani2017attention,brundage2018malicious,bender2021dangers,cath2018governing}.

In response to these risks, a multi-layered ecosystem of risk, incident and hazard reporting mechanisms has emerged, encompassing community-driven platforms, crowdsourcing initiatives, and expert-led assessments. Community-driven platforms, such as Reddit and Discord, allow users to collectively identify and flag potential issues in real time, offering diverse perspectives on emerging risks. Crowdsourcing initiatives, like those featured in DEF CON's Generative Red Team AI and OpenAI's Preparedness Challenge, mobilise large groups of individuals to rigorously test AI systems for vulnerabilities. For instance, the AI Village at DEF CON 31 hosted the largest-ever public Generative AI Red Team event, engaging over 2,000 participants to evaluate models from leading AI organisations 
\cite{HumaneIntelligence2024}. Similarly, OpenAI's Preparedness Challenge invited the public to identify potential risks in AI systems, offering incentives such as API credits for top submissions \cite{openai2024preparedness}. These initiatives not only enhance the robustness of AI models but also foster a collaborative approach to AI safety, integrating insights from diverse stakeholders to proactively address potential threats.

We hypothesise that supervisory entities will need to increase their focus on AI risk management as the volume of risk reports on GPAI models continues to grow. This influx of reports could, over time, challenge the capacity of these entities to address them comprehensively, making the development of efficient strategies for prioritising and addressing risks essential. A comparable situation has been documented 
 in the context of General Data Protection Regulation (GDPR) enforcement in the European Union.
 The Irish Council for Civil Liberties has reported on the cost and work load linked to GDPR enforcement \cite{iccl2023gdpr}. Dreschsler signals that \emph{the empowerment of individuals is thus to an extent stunted by the ability of the enforcement structure to cope with the requests} \cite{drechsler2022sword}.
  While the circumstances surrounding GDPR enforcement are varied and nuanced, this example helps to illustrate the kinds of challenges that supervisory authorities might encounter as they work to ensure comprehensive oversight of growing AI-related risks.

\begin{figure*}[htbp]
    \centering
    \includegraphics[width=0.6\textwidth]{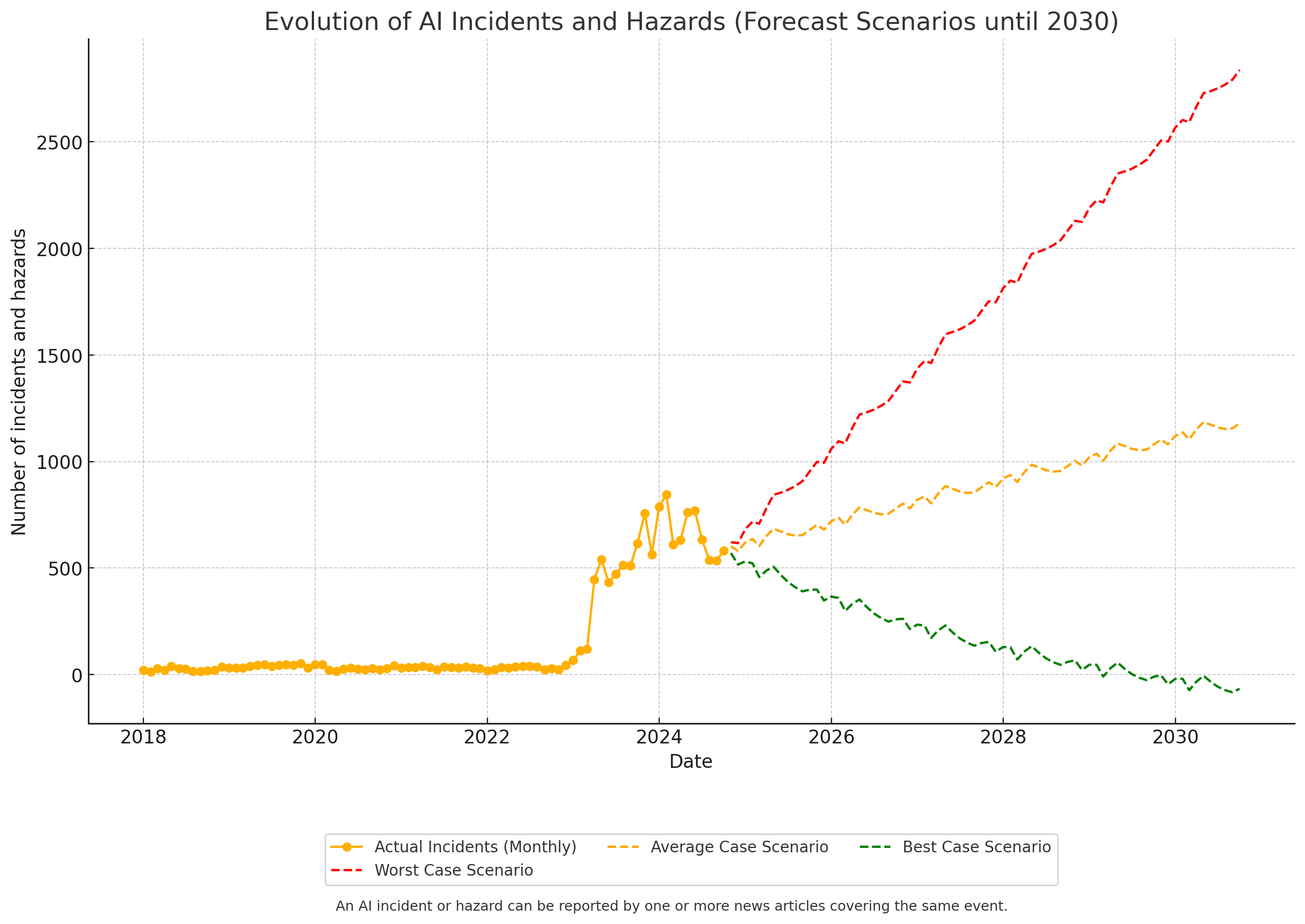}
    \includegraphics[width=1\textwidth]{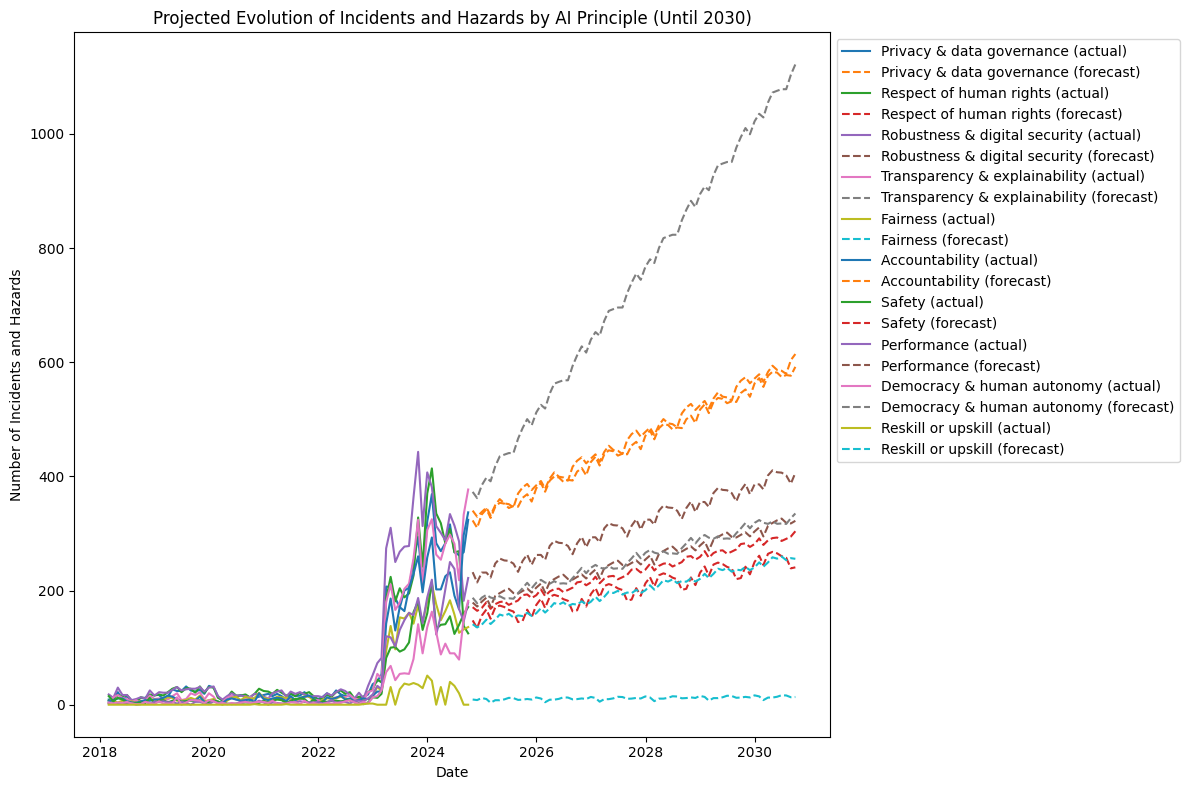}
    \caption{(Top) Scenario Projections for Global Evolution of AI Incidents and Hazards (Bottom) Projected Evolution of Incidents and Hazards by AI Principle.}
    \label{fig:combined_projections}
\end{figure*}

Our study examines how supervisory entities can process and prioritise these incoming AI risk reports. In the context of this paper, the terms 'risk' and 'risk reporting' are used broadly to encompass not only potential risks, but also incidents and hazards \cite{d1a8d965-en} associated with GPAI models.
As the number of reports likely exceeds the capacity for thorough investigation and mitigation, the policies guiding the selection and prioritisation of risks will significantly influence the trajectory of AI safety outcomes. To explore these challenges, we develop a simulation framework to model the generation and processing of GPAI risk reports from various sources—community platforms, crowdsourced initiatives, and expert analyses. By simulating different supervision policies—ranging from non-prioritised, first-come-first-served approaches to priority-based and diversity-prioritised strategies—we aim to uncover how these policies impact risk mitigation efficiency and the handling of diverse risk types. Additionally, we incorporate feedback loops between reporting incentives and risk deterrence to explore how prioritising expert-driven reports might skew the risk landscape by discouraging community input.

In alignment with the EU AI Act \cite{AIA24}, we contextualise our work within the broader category of GPAI models, as defined by the Act's Article 3(63): \emph{“an AI model, including where such an AI model is trained with a large amount of data using self-supervision at scale, that displays significant generality and is capable of competently performing a wide range of distinct tasks regardless of the way the model is placed on the market and that can be integrated into a variety of downstream systems or applications”}. This definition underscores the supervision challenges posed not only by text-based LLMs but also by multimodal systems capable of processing and generating diverse data types.

Our simulation framework is designed to support a wide range of supervisory bodies, including national AI market surveillance authorities, non-governmental organisations, corporate AI ethics boards, and industry consortia. Each of these entities must navigate the challenges of processing and prioritising large volumes of AI risk reports from diverse sources \cite{critch2020ai, hendrycks2021unsolved, bletch2023summit}.

To validate our framework, we applied it to real-world data from the Allen Institute for AI's WildChat dataset, which contains over one million conversations between users and ChatGPT, including more than 150K identified as containing toxic content. This dataset of real-world interactions provides an empirical basis to test the effectiveness of different prioritisation strategies and their practical implications for managing risks in GPAI models \cite{zhao2024wildchat}.

In the following section, we underscore the urgency for proactive supervisory frameworks by projecting the future trajectory of AI incidents and hazards. These forecasts establish a basis for the policy evaluations and risk management strategies discussed throughout the remainder of the paper. We then outline our methodology, present a comparative analysis of four supervision policies, and explore the broader implications of our findings for AI governance and safety.

\section{Forecasting Future AI Incidents}

To project the trajectory of AI incidents, we analysed data from the OECD AI Incidents Monitor (AIM) \cite{oecd_ai_incidents}, 
which covers AI incidents and hazards from January 2018 to October 2024, including incidents caused by GPAI models as well as other types of AI systems or AI application contexts. Initial incident levels averaged approximately 20 per month in 2018, rising sharply 
in parallel to the proliferation of GPAI models.
By late 2023, monthly incidents exceeded 500, with peaks reaching 844 in early 2024, underscoring the rapid escalation of AI-related risks alongside intensified deployment (the code used for these projections is available in the Code and Data Availability section). 

Using exponential smoothing, we developed three scenarios to forecast incidents through 2030 (Figure~\ref{fig:combined_projections}, top). In the worst-case scenario, unchecked AI adoption could lead to exponential growth, with incidents potentially surpassing 2,500 per month by 2030, posing significant regulatory challenges. The average-case scenario suggests moderate growth, with incidents reaching around 1,000 per month by 2030, indicating that regulation can slow growth but may not fully contain it. The best-case scenario projects a decline in incidents post-2025 under strong regulatory oversight, potentially reducing incidents below current levels by 2030. These figures correspond to global incidents, so the values for a specific AI supervisory entity should be adapted to the specific jurisdictional scope of that entity.

In addition to overall incident rates, we examined the projected evolution across key AI principles, such as Privacy, Robustness, Transparency, and Fairness (Figure~\ref{fig:combined_projections}, bottom). Incidents related to Robustness \& Digital Security are expected to see the sharpest increase, potentially exceeding 1,000 monthly cases by 2030, underscoring significant security concerns. Incidents associated with Transparency \& Explainability and Accountability are also projected to grow steadily, highlighting the need for governance frameworks that prioritise these principles. Meanwhile, issues related to Privacy and Human Rights are projected to grow moderately, suggesting ongoing but contained challenges as AI systems integrate with sensitive societal contexts.

To further understand the dynamics of AI risks, we examined the evolution of jailbreak prompts, a specific category of adversarial prompts used to bypass LLM safeguards. Our analysis is based on data collected by Shen et al. \cite{shen2023jailbreak}, which includes over 1,400 jailbreak prompts spanning December 2022 to December 2023. The study reveals that these prompts, originating from various online communities, have grown in sophistication and frequency, often exploiting vulnerabilities through prompt injection and privilege escalation techniques.

\begin{figure}[t!]  
    \centering
    \includegraphics[width=0.5\textwidth]{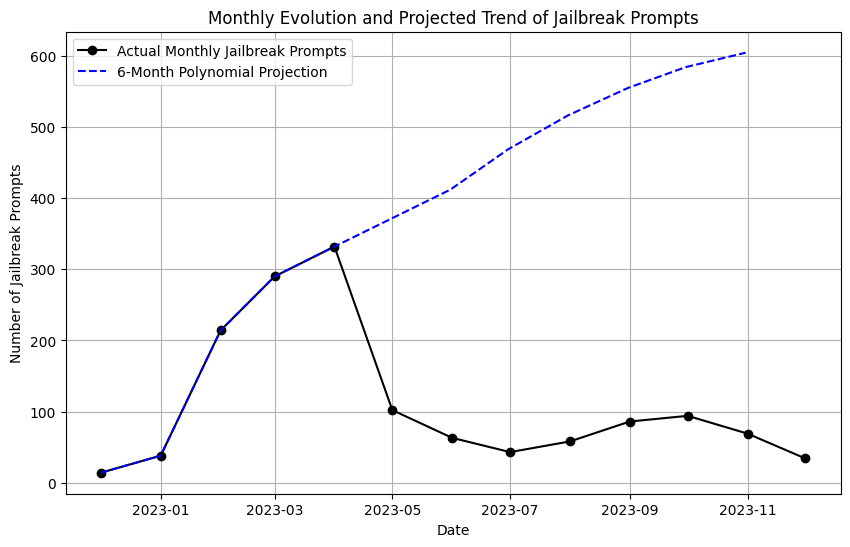} 
    \caption{Projected Evolution of Jailbreak Prompts with and without the June 2023 Regulatory Intervention.} 
    \label{fig:jailbreaks} 
\end{figure}

A pivotal event occurred in June 2023, when an update to OpenAI's ChatGPT made it more challenging for users to execute jailbreak prompts \cite{zhao2024wildchat}. This intervention can be seen as a form of corporate regulatory action, analogous to how policy-driven regulations might constrain AI misuse. Figure~\ref{fig:jailbreaks} shows two scenarios: one projecting the upward trajectory of jailbreak prompts if no intervention had occurred and another reflecting the observed decrease after the June update. Without the intervention, jailbreaks might have continued to proliferate, illustrating the tangible impact that regulatory measures—whether corporate or governmental—can have on curbing AI misuse.

These projections, segmented by scenario and principle, provide critical insights into the areas that may require immediate regulatory focus and those that could benefit from sustained oversight. They illustrate the diverse risk landscape associated with GPAI models and underscore the need for targeted, proactive policy responses—setting the stage for the policy evaluations and simulations in the following sections.

\section{Simulation framework}

Our study employs a simulation framework to compare four supervision policies for processing GPAI models risk reports: (1) \textit{non-prioritised}, (2) \textit{random}, (3) \textit{priority-based}, and (4) \textit{diversity-prioritised}. The \textit{non-prioritised} policy serves as a baseline, representing a first-come, first-served approach. While simple to implement, it may not efficiently address critical risks. 

The \textit{random} policy, which processes reports in a randomised order, serves as an important control condition. This approach helps to isolate the effects of structured prioritisation strategies by providing a comparison point that avoids systematic biases while still differing from the chronological processing of the \textit{non-prioritised} policy. The \textit{random} policy can be particularly useful in assessing whether observed differences in outcomes are due to intentional prioritisation or simply the result of processing reports in a non-chronological order.

The \textit{priority-based} policy, inspired by traditional risk management practices, strategically addresses the most significant risks (in terms of accessibility or damage) first.

The \textit{diversity-prioritised} policy aims to balance prioritisation with comprehensive coverage across different risk types, addressing the interconnected nature of GPAI risks and preventing potential blind spots in mitigation efforts \cite{hendrycks2021unsolved, weidinger2021ethical, liang2023holistic}.

Full details of the simulation framework, including parameter specifications and processing algorithms, are available in the Code and Data Availability section. The parameters were carefully chosen to qualitatively resemble the characteristics observed in prior studies on community, crowdsourced, and expert reporting.

\subsection{Report Prioritisation}

We define a set of reports \( R = \{r_1, r_2, \ldots, r_n\} \) generated over the simulation period, where each report \( r_i \) captures essential attributes relevant to AI risk supervision.

Each report has an arrival time \( t_i \), indicating when the risk report becomes available for review. The supervision cost \( s_i \) represents the resources (e.g., time, expertise, computational power) needed to assess and address the reported risk adequately, with higher costs reflecting more complex risks requiring specialised attention.

Accessibility \( a_i \) reflects how easily a reported risk could be encountered or exploited, with values ranging from 0 to 1 \cite{burden2024conversational}. Accessibility values differ by source, following distinct Beta distributions that characterise the accessibility patterns of each source. Community reports typically yield higher accessibility values, as they are generated from a \(\text{Beta}(5, 2)\) distribution, indicating issues frequently encountered by general users. Crowdsourced reports, with a \(\text{Beta}(3, 3)\) distribution, provide a more balanced accessibility range, while expert reports, following a \(\text{Beta}(2, 5)\) distribution, tend to have lower accessibility values, reflecting more specialised, less commonly encountered risks.

The potential damage \( d_i \), which quantifies the severity of the negative impact if the risk were to materialise, has a broader range due to its distribution and scaling based on the source. Potential damage is modelled using a Pareto distribution, with different scaling factors to capture the expected impact levels from each source. Community reports are scaled by 100, representing lower typical damage, while crowdsourced reports are scaled by 200, allowing for a broader potential damage spectrum. Expert reports, scaled by 500, reflect the higher potential for serious consequences in risks identified by experts, emphasizing their systemic importance even if encountered less frequently.  The Pareto distribution for potential damage mirrors the long-tailed severity profiles noted in large-scale red-teaming and crowdsourcing initiatives \cite{perez2022red,ganguli2022red}, where a small fraction of discovered issues accounts for disproportionately large risks.

Together, accessibility and potential damage inform the prioritisation of each report, as captured by the priority score \( p_i \), calculated as:
\begin{equation}
p_i = \log(1 + a_i \cdot d_i)
\end{equation}

This scoring function combines accessibility and potential damage, giving prominence to high-accessibility, high-damage risks. The logarithmic scaling tempers the influence of extreme values, ensuring that while high-risk cases are prioritised, they do not disproportionately dominate the scoring, as illustrated in Figure \ref{fig:priority_illustration} in the supplementary information.

Each report also includes a source attribute, \( src_i \in \{\text{com, crd, exp}\} \), denoting whether the report originates from a community member, a crowdsourcing initiative, or an expert.

Finally, each report includes a risk type \( rt_i \), which enables organised analysis of the kinds of risks emerging in the AI ecosystem. Broadly, risk types in our framework include privacy risks, security vulnerabilities, bias, ethical considerations (including AI alignment), robustness, and content moderation. Privacy risks involve potential breaches or misuse of personal data, while security risks focus on vulnerabilities that could disrupt AI systems. Bias risks highlight concerns about unfair or discriminatory outcomes, while ethical risks address alignment with societal norms and long-term societal impact. Robustness concerns the resilience of the system, and content moderation addresses the risk of generating harmful or misleading content. This risk categorisation is not intended to be comprehensive or exemplary. It has been defined as a prerequisite to validate the proposed simulated framework, but other categorisations are possible.

\subsection{Report Generation}
Report generation is modeled as a Poisson process for each source:
\begin{equation}
N_{src}(t) \sim \text{Poisson}(\lambda_{src} \cdot t)
\end{equation}
where the parameters were empirically set to $\lambda_{com} = 25$, $\lambda_{crd} = 12$, and $\lambda_{exp} = 5$ per month as explained below (an illustrative figure showing the Poisson distributions for each source type is available in Figure \ref{fig:report_generation_distribution} in the supplementary information). 
We chose a Poisson process as we assume events (in this case, reports) occur independently of each other, which is a reasonable assumption for diverse sources of GPAI model risk reports.  The specific rates ($\lambda_{src}$) for each source were chosen based on careful consideration of the nature of each reporting channel. For community sources, we set a high rate ($\lambda_{com} = 25$) to reflect the large number of users of GPAI models and the ease of submitting reports through platforms like social media or dedicated forums \cite{sun2021safety,bhardwaj2023red,shen2023jailbreak}. This high volume is expected to cover a wide range of issues from user experience to observed biases.

Crowdsourced reports are assigned a moderate rate ($\lambda_{crd} = 12$), representing organised efforts to identify risks but with fewer participants than the general community \cite{perez2022red,ganguli2022red}. These reports are expected to be more focused than community reports but more numerous than expert analyses.

For expert sources, we assume a lower rate ($\lambda_{exp} = 5$), considering the limited number of AI safety experts and the time-intensive nature of their analyses \cite{shi2023red,gopal2023releasing}. While fewer in number, these reports are likely to be more comprehensive and to address more complex, systemic risks \cite{cyberseceval2023,soice2023democratize}.

\subsection{Parameter Distributions}

The supervision cost \( s_i \) is modeled using a log-normal distribution because it naturally captures the right-skewed nature of resource requirements typically seen in risk assessment tasks. The parameters are set to reflect increasing complexity from community to expert sources. Community reports (\(\mu_{\text{com}} = 1.5, \sigma_{\text{com}} = 0.5\)) are assumed to require less time and expertise to process, while expert reports (\(\mu_{\text{exp}} = 3.0, \sigma_{\text{exp}} = 0.7\)) are more resource-intensive due to their typically more complex and technical nature. Crowdsourced reports fall in between (\(\mu_{\text{crd}} = 2.0, \sigma_{\text{crd}} = 0.6\)).

Accessibility \( a_i \) is modeled using a beta distribution. Community reports are given parameters (\(\alpha_{\text{com}} = 5, \beta_{\text{com}} = 2\)) that skew towards higher accessibility, reflecting the idea that issues reported by general users are often more readily observable or exploitable. Expert reports, with parameters (\(\alpha_{\text{exp}} = 2, \beta_{\text{exp}} = 5\)), tend towards lower accessibility, representing more obscure or complex vulnerabilities that might be harder to exploit but potentially more serious. Crowdsourced reports (\(\alpha_{\text{crd}} = 3, \beta_{\text{crd}} = 3\)) are modeled with a symmetric distribution, representing a balance between the two extremes. The Beta distributions chosen to represent accessibility reflect skewed patterns observed in community-driven vulnerability identification efforts, such as those studied by Shen et al. \cite{shen2023jailbreak}, where frequently reported exploits cluster in more accessible regions.

Potential damage \( d_i \) follows a Pareto distribution, which is commonly used to model the distribution of large events in complex systems. This choice reflects the “long-tail” nature of AI risks, where most issues might have relatively low impact, but a few could have extreme consequences. The parameters are set to model increasing potential damage from community to expert sources. Community reports have the highest shape parameter (\(a_{\text{com}} = 3\)) and lowest scale (\(k_{\text{com}} = 100\)), representing mostly lower-impact issues with fewer extreme cases \cite{sun2021safety}. Expert reports have the lowest shape parameter (\(a_{\text{exp}} = 1.5\)) and highest scale (\(k_{\text{exp}} = 500\)), modeling a higher likelihood of identifying severe, high-impact risks \cite{alfonseca2005common}. Crowdsourced reports again fall in the middle (\(a_{\text{crd}} = 2, k_{\text{crd}} = 200\)).
An illustrative figure showing the distributions of supervision cost, accessibility, and potential damage for each source type is available in the supplementary information (Figure \ref{fig:parameter_illustration}).

\subsection{Source-Based Risk Type Assignment and Distribution}

In our framework, risk type assignment is closely tied to the report source, capturing the different perspectives and priorities of community users, crowdsourced contributors, and expert assessors. For community sources, we assign higher probabilities to risk types likely to be encountered and recognised by general users. For example, issues related to ``user experience" have a high probability (0.3) for community-sourced reports, reflecting the tendency of everyday users to notice and report problems that directly impact their interactions with GPAI models, such as interface difficulties, unexpected model responses, or perceived biases in outputs \cite{sun2021safety}.

In contrast, crowdsourced reports are given a more balanced distribution across risk types but with slightly higher probabilities for categories that align with common concerns within the AI ethics community \cite{perez2022red}. This balance reflects the nature of crowdsourcing efforts, which typically involve individuals with varying levels of expertise and specific interests in AI safety \cite{ganguli2022red}. This distribution approach recognises that crowdsourced contributors may bring a mixture of general observations and more targeted insights into AI-related risks.

Expert sources, on the other hand, are assigned higher probabilities for technical and long-term risk categories. For instance, the probability that an expert report will address AI alignment issues is set to 0.2, capturing the tendency of AI researchers and ethicists to focus on fundamental challenges related to ensuring AI behavior aligns with human values and intentions \cite{gabriel2020artificial}. This higher likelihood of focusing on alignment and other deep technical issues underscores the unique insights that experts bring to understanding complex, systemic AI risks. 

\begin{table}[H]
\centering
\caption{Source-specific prior probabilities for each risk type.}
\label{tab:prior_risk_types}
\footnotesize            
\setlength{\tabcolsep}{4pt}   
\begin{tabularx}{\columnwidth}{l *{3}{>{\centering\arraybackslash}X}}
\toprule
\textbf{Risk type} & \tiny{Community} & \tiny{Crowdsourced} & \tiny{Expert} \\
\midrule
Privacy                       & 0.30 & 0.20 & 0.00 \\
Misinformation                & 0.25 & 0.20 & 0.00 \\
Bias                          & 0.20 & 0.15 & 0.00 \\
User experience               & 0.15 & 0.00 & 0.00 \\
Content moderation            & 0.10 & 0.00 & 0.00 \\
Security                      & 0.00 & 0.15 & 0.20 \\
Ethical                       & 0.00 & 0.15 & 0.20 \\
Robustness                    & 0.00 & 0.15 & 0.15 \\
Long-term societal impact     & 0.00 & 0.00 & 0.20 \\
AI alignment                  & 0.00 & 0.00 & 0.15 \\
Interpretability              & 0.00 & 0.00 & 0.10 \\
\bottomrule
\end{tabularx}
\label{tab:prior_risk_types}
\end{table}

The risk types in our framework cover key areas vital to comprehensive AI risk management. Privacy risks address concerns around misuse or unauthorised exposure of personal data in interactions with the GPAI model. Security risks focus on vulnerabilities that could disrupt or compromise AI systems. Bias risks involve potential unfair or discriminatory outputs from language models, which may mirror societal biases. Ethical and alignment risks emphasise ensuring AI behaviour aligns with human values and goals \cite{gabriel2020artificial}. Robustness evaluates the system’s resilience across various conditions, while content moderation and misinformation risks concern the generation of harmful or misleading content. Each risk type has a source-specific distribution, supporting a realistic simulation of community, crowdsourced, and expert reporting patterns. Full details of the risk-type assignment probabilities by source are
summarised in Table~\ref{tab:prior_risk_types} and are also available
in the simulation code, as noted in the Code and Data Availability section.

\subsection{Simulation Parameters}

The simulation parameters are designed to model the operations of an AI supervisory entity over an extended period. During each time step \( t \), which corresponds to a month, the entity processes reports generated by various sources. The total simulation duration is \( T \), with an initial observation period \( T_{\text{obs}} \). During this observation period, the entity collects data to understand the volume, complexity, and nature of incoming reports. This initial phase helps the entity calibrate its processing and resource allocation strategies.

The processing capacity per month, \( C_m \), is calculated based on the average total supervision cost of the reports received during the observation period. The reports received in each month of the observation period are denoted as \( R_t \), where \( R_t \) is the set of reports generated at time \( t \). Therefore, \( C_m \) is defined as:

\begin{equation}
C_{m} = C_0 \cdot \frac{1}{T_{\text{obs}}} \sum_{t=1}^{T_{\text{obs}}} \sum_{r_i \in R_t} s_i
\end{equation}

Secondly, setting \( C_0 < 1 \) creates a challenging scenario where efficient prioritisation becomes crucial for managing the inevitable backlog, mirroring the resource pressures often faced by regulatory bodies. This scenario is reminiscent of the backlog issues encountered in the implementation of the General Data Protection Regulation (GDPR), where regulatory agencies frequently face limited resources relative to the volume of compliance reports and complaints \cite{iccl2023gdpr, drechsler2022sword}. Such constraints necessitate efficient prioritisation to address high-impact cases effectively.

\subsection{Supervision Policies}
We formalise the four distinct supervision policies:\\

\textit{Non-Prioritised} (NP):
\begin{equation}
P_t^{NP} = \{r_i \in R_t : \sum_{r_j \in P_t^{NP}} s_j \leq C_m\}
\end{equation}

\textit{Random} (RD): \begin{equation} P_t^{RD} = \text{RS}(R_t, n_t), n_t = \max{n : \sum_{i=1}^n s_{(i)} \leq C_m} \end{equation}

\textit{Priority-based} (PB):
\begin{equation}
P_t^{PB} = \underset{S\subseteq R_t, \sum_S s_i \leq C_m}{\text{arg max}}\sum_S p_i
\end{equation}

\textit{Diversity-prioritised} (DP):
\begin{equation}
P_t^{DP} = \underset{S\subseteq R_t, \sum_S s_i \leq C_m}{\text{arg max}}\sum_S p_i \delta_i(S)
\end{equation}

Here, $\delta_i(S)$ is a diversity score for report $r_i$ given set $S$. $R_t$ is the set of available reports at time $t$, \text{RS} denotes a random sampling operation., $s_i$ is the supervision cost of report $r_i$, $p_i$ is its priority score, and $C_m$ is the monthly processing capacity.

To clarify how processing reports leads to tangible mitigation, we assume a concrete intervention mechanism tied to each source and risk type. For example, when an expert-sourced report regarding a severe model vulnerability is processed, it triggers a targeted mitigation action—such as deploying a model patch, revising content filtering parameters, or refining detection rules for harmful outputs. Similarly, addressing community-reported issues might result in improved user-facing safeguards, and crowdsourced reports could inform updates to red-teaming protocols. These interventions directly reduce the future likelihood of similar incidents. As a result, the cumulative number and type of processed reports—not just their existence—translate into measurable improvements in model resilience and a quantifiable reduction in long-term hazards. This explicit causal link between processing decisions and outcome improvements ensures that prioritization strategies have a clear, practical impact on the evolving risk landscape.

\section{Illustration of a No-Prioritisation Run}

\begin{figure*}[htbp]
    \centering
    \includegraphics[width=0.6\textwidth]{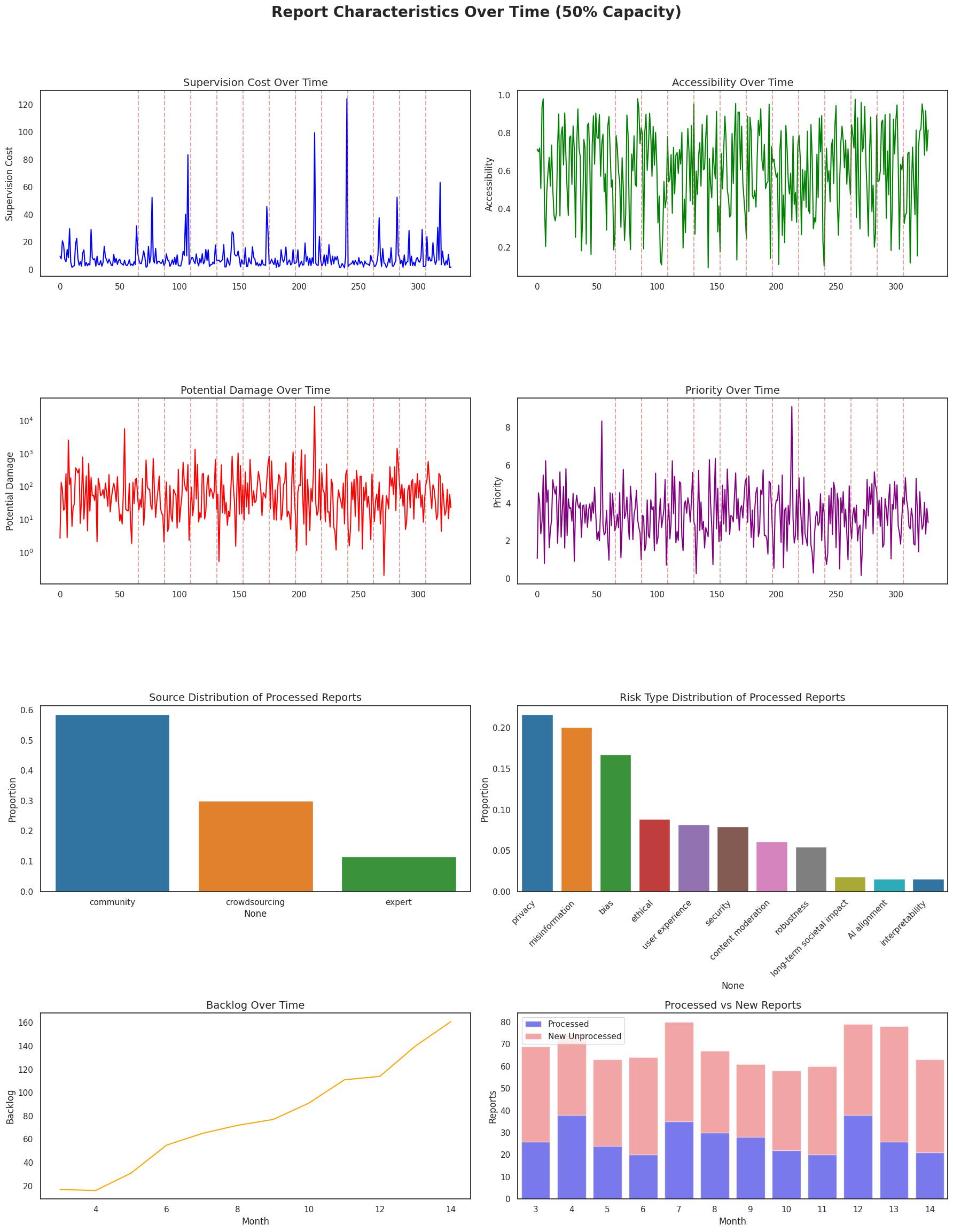}
    \caption{Results of a typical run using the no-prioritisation policy for LLM risk report processing. Proportions are calculated based on the total number of processed reports in the simulation.}
    \label{fig:no_prioritization_run}
\end{figure*}

To better understand the dynamics of GPAI model risk report processing under different policies, we first examine a typical run using the \textit{no-prioritisation} approach. This policy processes reports in the order they are received, without considering their potential impact or urgency. Figure \ref{fig:no_prioritization_run} presents a view of the simulation results over a 15-month period, with the first 3 months serving as an observation period to calibrate the monthly processing capacity.

The top four graphs in Figure \ref{fig:no_prioritization_run} illustrate the characteristics of processed reports over time. The Supervision Cost graph shows significant variability, with occasional spikes indicating reports that require substantial resources to process. This variability highlights the challenges in resource allocation when reports are processed without prioritisation. The Accessibility and Potential Damage graphs similarly display considerable fluctuations, suggesting that high-impact or easily exploitable risks are not consistently addressed in a timely manner under this policy.

The Priority graph, which combines Accessibility and Potential Damage, further emphasises the lack of strategic processing. High-priority reports (indicated by peaks in the graph) are interspersed with lower-priority ones, reflecting the chronological processing approach of the no-prioritisation policy.

The fifth and sixth graphs depict the distribution of report sources and risk types for the processed reports. Under the no-prioritisation policy, a higher proportion of reports originate from the community, followed by crowdsourced reports, with expert reports being the least processed. In terms of risk type, privacy and misinformation issues dominate, while more complex or long-term risks, such as AI alignment or interpretability, receive relatively less attention.

The Backlog graph shows a steady increase in unprocessed reports over time. This growing backlog indicates that the no-prioritisation policy struggles to keep pace with the influx of new reports, potentially leading to delays in addressing important risks.

The final graph compares the number of processed reports to new unprocessed reports each month. The consistent gap between these two categories illustrates the ongoing challenge of managing report volume within resource constraints. 

\section{Illustration of a Priority-Based Run}

\begin{figure*}[htbp]
    \centering
    \includegraphics[width=0.6\textwidth]{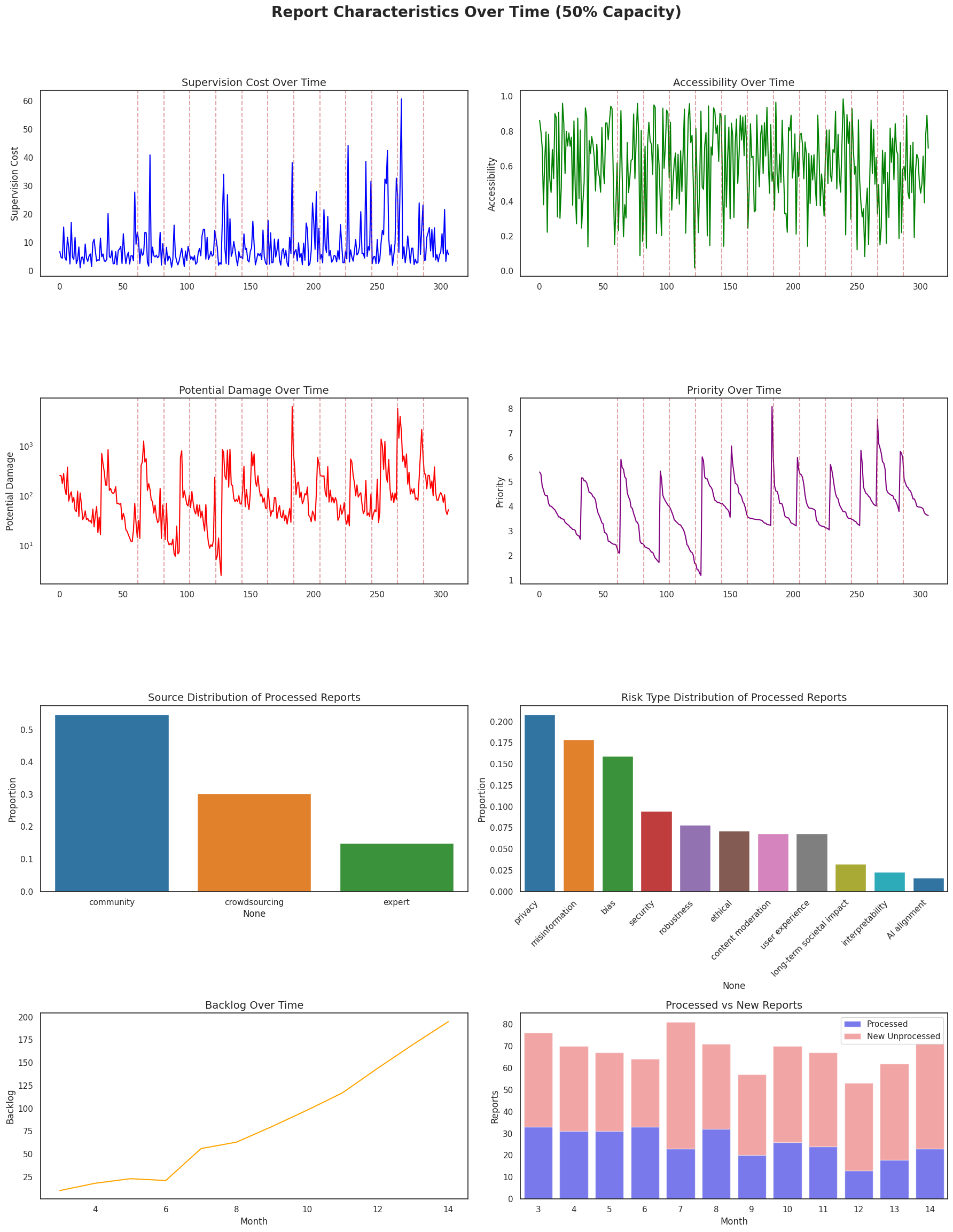}
    \caption{Results of a typical run using the priority-based policy for LLM risk report processing. Proportions are calculated based on the total number of processed reports in the simulation.}
    \label{fig:priority_based_run}
\end{figure*}

To contrast with the \textit{non-prioritised} approach, we now examine a typical run using the \textit{priority-based} policy for GPAI model risk report processing. This policy selects reports based on their calculated priority scores, aiming to address the most critical risks first. Figure \ref{fig:priority_based_run} presents the simulation results over the same 15-month period as the \textit{non-prioritised} example.

The top four graphs in Figure \ref{fig:priority_based_run} reveal several key differences. First, the Supervision Cost shows a broader range of values with more frequent spikes compared to the \textit{non-prioritised} run. This indicates that the \textit{priority-based} policy often processes reports with higher supervision costs, likely focusing on more complex issues.

In the Potential Damage graph, there is a noticeable trend towards processing reports with higher potential damage, which is less apparent in the \textit{non-prioritised} run. This reflects the policy's effectiveness in prioritizing high-risk reports, addressing them earlier and more consistently.

The Priority graph also shows a stark difference, as the \textit{priority-based} approach consistently focuses on higher-priority reports throughout the simulation. Unlike the \textit{non-prioritised} run, where reports were processed based on arrival time, here we see a clear bias towards high-priority items, with very few low-priority reports being processed.

The Source Distribution remains similar, but there is a slight increase in the proportion of expert and crowdsourcing reports being processed, compared to the dominance of community reports in the \textit{non-prioritised} run. This shift suggests that the \textit{priority-based} policy gives more weight to reports from sources that may identify higher-priority risks.

\section{Comparison across prioritisation policies}

We present results from one batch of 100 simulations for each of the four policies: \textit{non-prioritised}, \textit{random}, \textit{priority-based}, and \textit{diversity-prioritised}. Each simulation was run with standard parameters: a 15-month duration (including a 3-month initial observation period) and a monthly processing capacity set to 50\% of the average total supervision cost observed during this period. Qualitatively similar results were observed in a second batch of simulations.

\begin{figure*}[htbp] \centering \includegraphics[width=\textwidth]{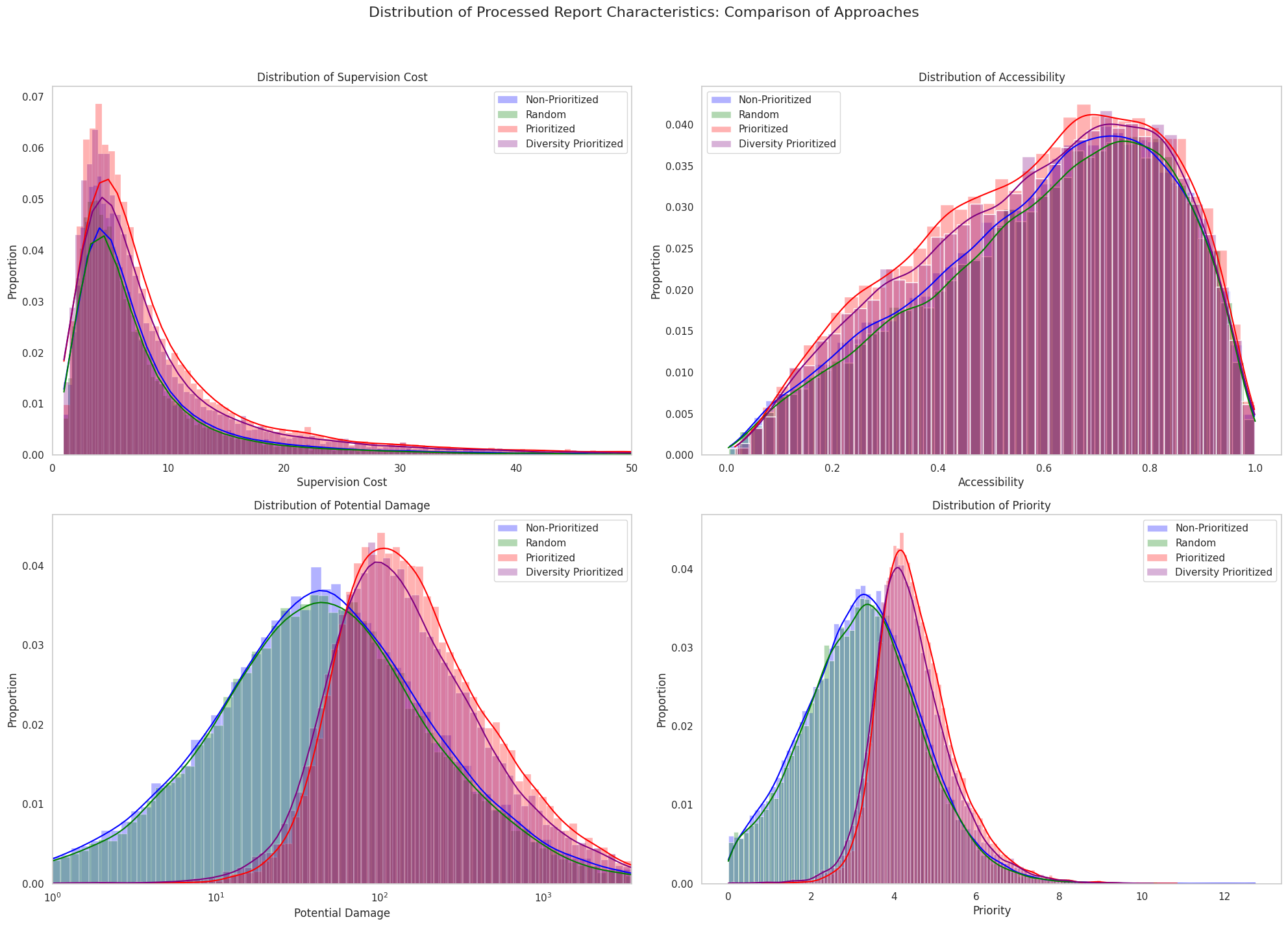} \caption{Distribution of Processed Report Characteristics: Comparison of Approaches.} \label{fig_report_} \end{figure*}

Figure \ref{fig_report_} illustrates the distribution of key report characteristics across the four policies. The figure consists of four histograms, each representing one of our key metrics: Supervision Cost, Accessibility, Potential Damage, and Priority.

The Supervision Cost histogram (top left) shows a right-skewed distribution for all policies, with the \textit{priority-based} and \textit{diversity-prioritised}approaches showing longer tails towards higher costs. The mean Supervision Cost was highest for the \textit{priority-based} approach (11.18), followed by \textit{diversity-prioritised} (9.67), \textit{non-prioritised} (8.60), and \textit{random} (8.17). The Kruskal-Wallis H-test confirmed a statistically significant difference in Supervision Cost across policies (H=54.42, p $<$ 0.01), indicating that prioritisation methods tend to select more complex or time-consuming reports.

In terms of Accessibility (top right), differences between policies were more subtle. The \textit{random} approach had the highest mean accessibility (0.6150), while the \textit{priority-based} approach had the lowest (0.5741). The Kruskal-Wallis H-test revealed a statistically significant difference across policies (H=15.94, p $<$ 0.01), although the variation is less pronounced than in other metrics.

The Potential Damage histogram (bottom left) reveals stark differences between policies. The \textit{priority-based} and \textit{diversity-prioritised} approaches focus more on reports with higher potential damage, with mean Potential Damage values of 558.78 and 457.86, respectively, compared to 181.48 for \textit{non-prioritised} and 205.43 for \textit{random}. The Kruskal-Wallis H-test (H=1103.53, p $<$ 0.01) shows this difference is statistically significant. Notably, the 90th and 99th percentiles for \textit{priority-based reports} show extreme potential damage values (1028.73 and 6212.36), suggesting that this policy effectively targets the most severe risks.

The Priority histogram (bottom right), combining Accessibility and Potential Damage, reflects similar patterns. The \textit{priority-based} (4.81) and \textit{diversity-prioritised} (4.54) approaches have the highest mean priority, significantly higher than \textit{non-prioritised} (3.25) and \textit{random} (3.22). The Kruskal-Wallis H-test (H=1245.19, p $<$ 0.01) indicates a significant difference across policies.

\begin{figure*}[htbp] \centering \includegraphics[width=\textwidth]{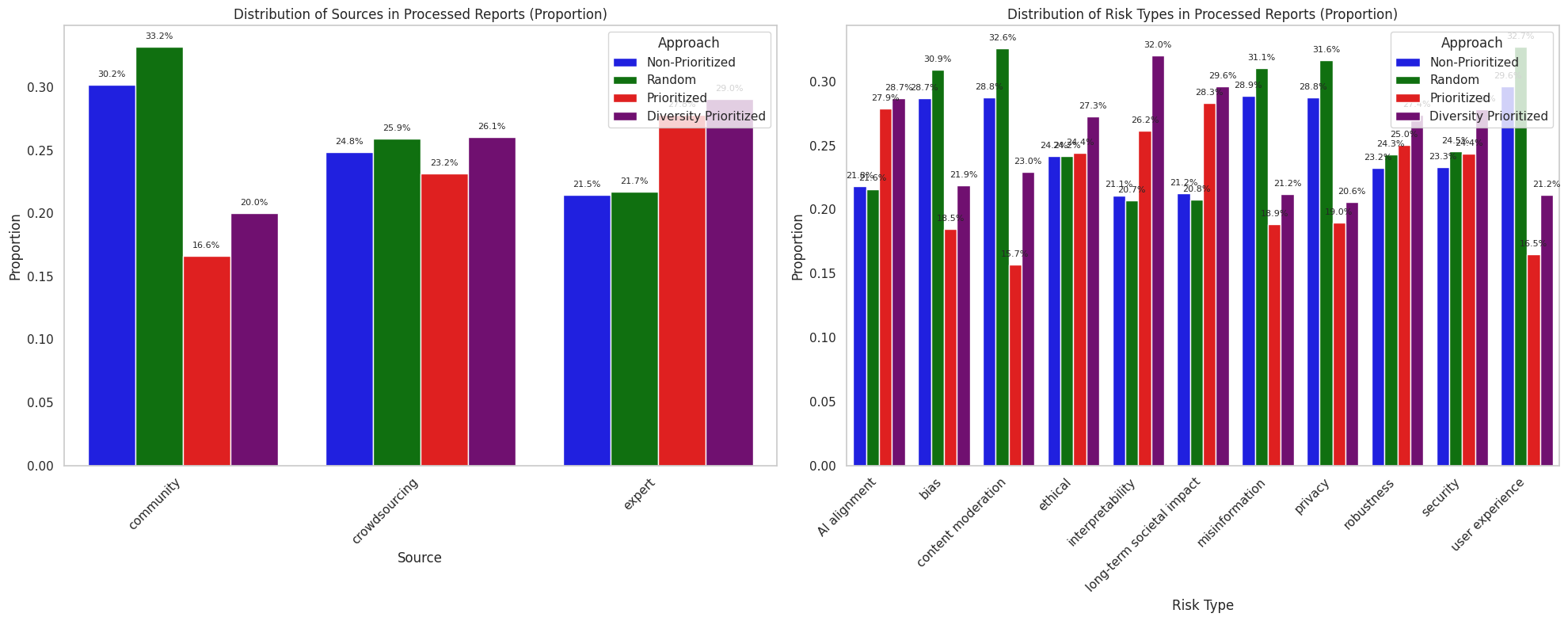} \caption{The left panel shows the proportion of reports processed from each source (community, crowdsourcing, expert) under different supervision policies. The right panel displays the distribution of processed risk types under each policy. Proportions in both panels are calculated relative to the total number of reports processed for each category. For example, in the left panel, the proportions represent the percentage of processed reports originating from each source under each policy. In the right panel, the proportions show the percentage of reports for each risk type that were processed under a given policy, relative to the total number of reports processed for that risk type.} \label{fig_source_distribution} \end{figure*}

Figure \ref{fig_source_distribution} provides insight into how different policies affect the distribution of report sources and risk types.

The left panel of Figure \ref{fig_source_distribution} shows the distribution of report sources (community, crowdsourcing, expert) for each policy. The \textit{non-prioritised} and \textit{random} approaches process a higher proportion of community reports, while the \textit{priority-based} and \textit{diversity-prioritised} approaches give more weight to expert and crowdsourced reports. This suggests that expert and crowdsourced reports tend to have higher priority scores, likely due to their potential for identifying more severe or complex risks.

The right panel displays the distribution of risk types across policies. Certain risk types, such as privacy, misinformation, and bias, are processed more frequently across all policies. However, the \textit{priority-based} and \textit{diversity-prioritised} approaches show a more even distribution across risk types, including greater attention to categories like ethical, security, and long-term societal impact. This indicates that these approaches may be more effective at addressing a diverse range of GPAI model risks, including those that may have more severe long-term consequences.

\section{Feedback Loops in Regulatory Prioritisation}

We can now model the feedback loops that influence the incentives for reporting and the deterrence of risks over time. These feedback loops help forecast the long-term prioritisation dynamics under different supervision policies, particularly the \textit{priority-based} and \textit{diversity-prioritised} approaches. These policies may unintentionally amplify expert reporting while under-addressing other sources and risks, driven by mechanisms of incentives and deterrence. For instance, most of the obligations of the EU AI Act focus on high-risk AI systems and GPAI models with systemic risks. This focus may result in increased reporting from experts in areas related to those systems and models, which could potentially lead to fewer resources being allocated to monitor lower-risk applications that might also pose significant concerns.

We start by modelling the incentives for each reporting source—community, crowdsourcing, and expert—which adjust based on the proportion of their reports that are processed. The incentive level for a source \( I_s(t) \) at time \( t \) evolves as follows:
\[
I_s(t+1) = \text{clip}\left( I_s(t) + \gamma_I \left( \pi_s(t) - \beta_I \right) \cdot I_s(t), I_{\text{min}}, I_{\text{max}} \right)
\]
where \( \pi_s(t) \) represents the processing rate of reports from source \( s \), \( \gamma_I \) controls how quickly incentives adjust, and \( \beta_I \) is the expected processing rate. The minimum and maximum incentive values, \( I_{\text{min}} \) and \( I_{\text{max}} \), constrain incentive levels within realistic bounds. Specifically, \( I_{\text{min}} = 0.5 \) ensures incentives do not drop to a discouraging level, while \( I_{\text{max}} = 1.5 \) prevents incentives from escalating unsustainably. Setting these bounds maintains a balanced reporting system by keeping incentives within manageable limits.

Simultaneously, the occurrence rate of each risk type \( O_{rt}(t) \) changes depending on how effectively it is addressed. Processed reports reduce the occurrence rate, reflecting both the enforcement effect of corrective measures and the deterrent effect of regulation:
\[
O_{rt}(t+1) = \max \left( O_{\text{min}}, O_{rt}(t) - \gamma_O \cdot M_{rt}(t) \right)
\]

However, risks that are not sufficiently addressed may resurface, with their occurrence rates recovering over time:
\[
O_{rt}(t+1) = \min \left( O_{\text{max}}, O_{rt}(t+1) + \delta_O \right)
\]

In these equations, \( \gamma_O \) represents the effectiveness of risk mitigation, \( M_{rt}(t) \) is the number of processed reports for risk type \( rt \), and \( \delta_O \) is the rate at which risks recover when not adequately managed. This model illustrates how regulatory focus directed toward high-priority risks, such as those identified by experts, can suppress these risks, while other less prioritised risks (e.g., user experience or content moderation) may re-emerge if not addressed.

\begin{figure*}[htbp]
    \centering
    \includegraphics[width=0.6\textwidth]{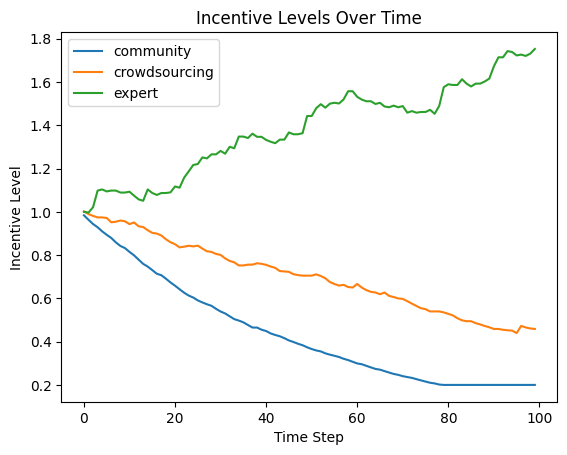}
    \includegraphics[width=0.6\textwidth]{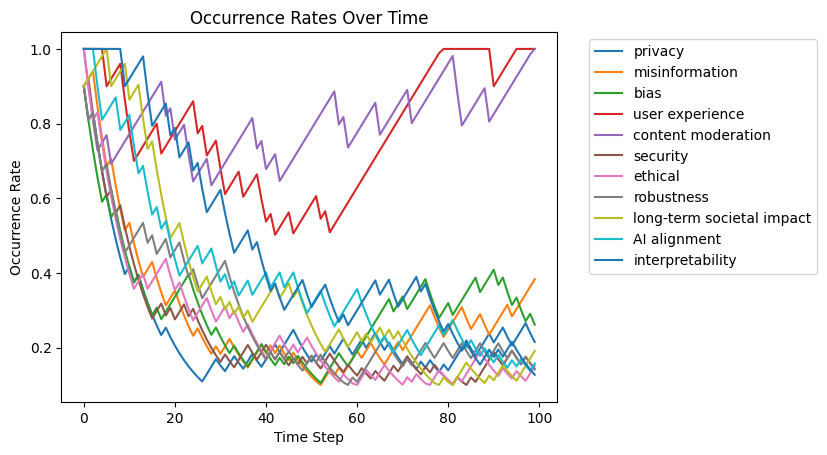}
    \caption{(Top) Incentive Levels Over Time: Expert incentives increase as attention shifts toward expert-driven reports, while community and crowdsourcing incentives decline. (Bottom) Occurrence Rates Over Time: High-priority risks such as security and AI alignment are suppressed, while user experience and content moderation re-emerge.}
    \label{fig:feedback_loops}
\end{figure*}

Figure \ref{fig:feedback_loops} illustrates these dynamics. The top plot shows how regulatory attention shifts incentives toward expert-driven reports, leading to increasing expert incentives over time while community and crowdsourcing incentives decline. The bottom plot demonstrates how high-priority risks, such as security and AI alignment, are successfully suppressed by regulatory actions, but less prioritised risks like user experience and content moderation tend to re-emerge as they receive less attention. This highlights the challenge of maintaining deterrence across all risk types and underscores the importance of a balanced approach to processing diverse types of risks, as exemplified by regulatory frameworks such as the General Data Protection Regulation (GDPR), which addresses both critical security risks and user-centric concerns.

\section*{Application to Real-World Data}

\begin{figure*}[htbp] 
    \centering 
    \includegraphics[width=0.6\textwidth]{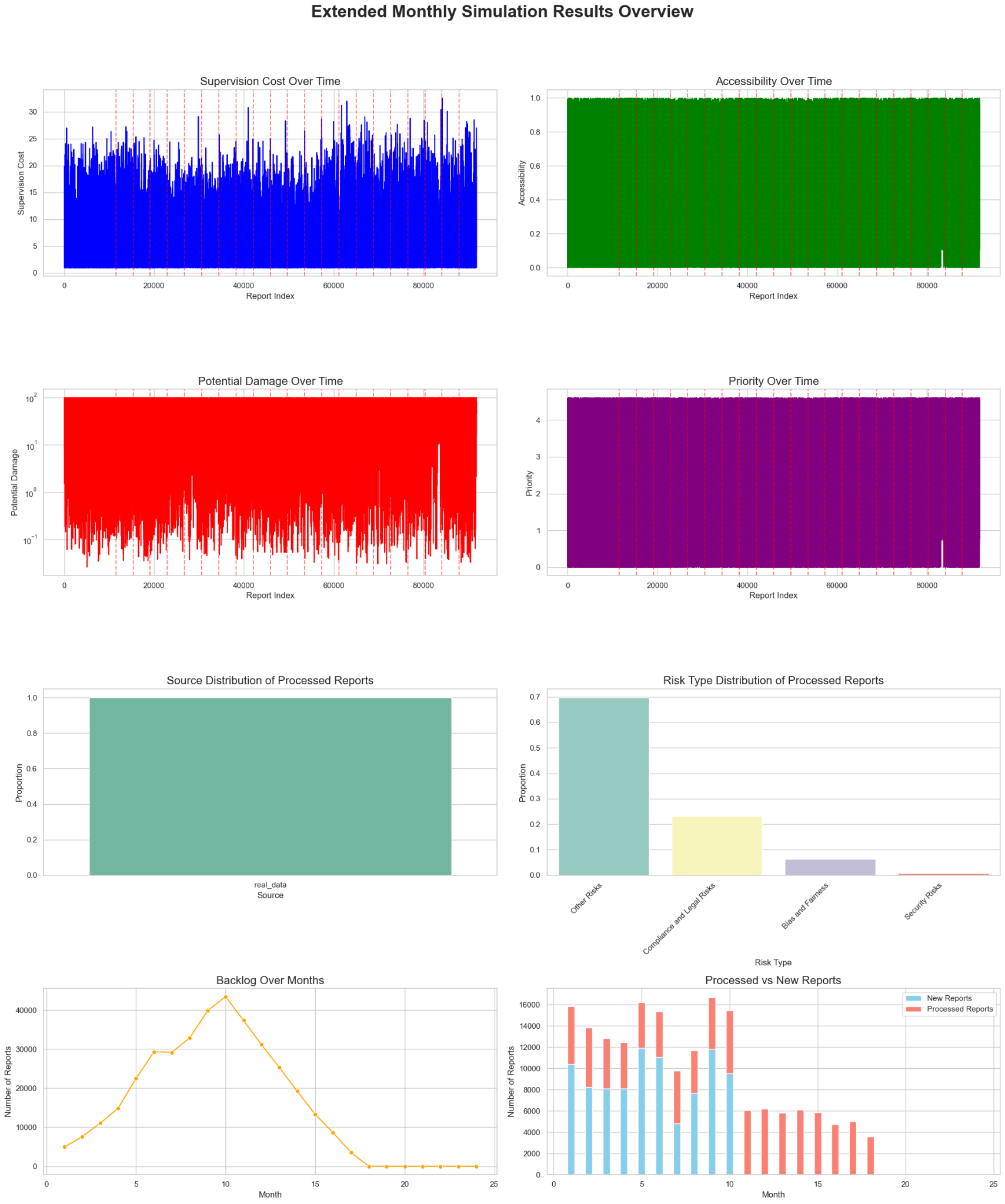} 
    \caption{Monthly Simulation Results Overview for the Non-prioritised Policy over the WildChat dataset. Simulation continues beyond the initial 12-month period to allow backlog processing to completion.} 
    \label{wildchat_non_prioritized} 
\end{figure*}

To validate our simulation framework and explore its practical implications, we applied our methodology to the WildChat dataset~\cite{zhao2024wildchat}, which comprises one million conversations between users and ChatGPT (versions 3.5 and 4). The dataset compilation spanned from April 9, 2023, to April 29, 2024, capturing over a year of real-world interactions with prominent large language models. WildChat includes more than 2.5 million interaction turns in 68 different languages, encompassing diverse user intents and interactions.

For this analysis, we focused on the subset of conversations identified as containing toxic content, totalling 152,296 conversations. These instances are particularly relevant for studying risk reporting and mitigation mechanisms in GPAI models, as they represent potential risks that supervisory entities need to manage.

We mapped the dataset's attributes to our simulation parameters—supervision cost ($s_i$), accessibility ($a_i$), potential damage ($d_i$), priority score ($p_i$), and risk type ($rt_i$)—as follows.

Supervision Cost ($s_i$), representing the resources required to assess and address a reported risk, was calculated based on aggregated toxicity scores from the Detoxify model~\cite{Detoxify}. We summed the scores for various toxicity categories (e.g. identity attack, insult, obscenity) for each conversation and applied a scaling factor to reflect the effort needed for mitigation. We ensured that $s_i$ was at least 1 to represent minimal supervision effort, and used a scaling factor $\kappa_s = 5$ to align with our previous simulations:

\begin{equation} 
s_i = \max\left(1, \kappa_s \sum_{\text{categories}} \text{score}_{i}\right). 
\end{equation}

Accessibility ($a_i$) is estimated from the number of dialogue turns
in each conversation.  Empirically, longer exchanges tend to involve
more iterative probing or prompt‐refinement, which in turn signals that
the underlying vulnerability is easier for attackers to
re-instantiate—a pattern also observed by \cite{burden2024conversational}.
To capture this monotonic relationship while keeping the metric in
\([0,1]\), we rescale turn count linearly and cap it at a saturation
point:

\begin{equation}
a_i = \min\!\Bigl(1,\; \frac{DT_i}{DT_{\max}}\Bigr),
\end{equation}

where \(DT_i\) is the number of dialogue turns in conversation \(i\).
The cap \(DT_{\max}=10\) corresponds to the \(90^{\text{th}}\)
percentile of turn counts in our WildChat sample; conversations longer
than ten exchanges are therefore treated as maximally accessible for
the purpose of the simulation.  This simple, data-anchored heuristic
avoids over-weighting the few outlier threads that continue well beyond
ten turns while still reflecting the positive correlation between turn
depth and exploit persistence.

Potential Damage ($d_i$) quantifies the negative impact if the reported risk materialises. We used the maximum Detoxify toxicity score across categories for each conversation,
   an approach previously shown to track annotator-rated harm levels
   in open-domain chats \cite{Detoxify,lin2023toxicchat}, and then scaled it
   by $\kappa_d  = 500$ to obtain potential damage:

\begin{equation} 
d_i = \kappa_d \max_{\text{categories}} \text{score}_{i}. 
\end{equation}

Priority Score ($p_i$) combines accessibility and potential damage, calculated as:

\begin{equation} 
p_i = \ln\left(1 + a_i d_i\right). 
\end{equation}

Risk Type ($rt_i$) was assigned based on detailed mappings from the toxicity categories provided by Detoxify. Each report was categorised by evaluating which toxicity scores exceeded specific thresholds, corresponding to predefined risk types. For example, if a conversation had a \texttt{threat} score above 0.5 or a \texttt{severe\_toxicity} score above 0.7, it was classified as a Security Risk. High scores in \texttt{sexual\_explicit} led to classification under Compliance and Legal Risks. Similarly, elevated scores in \texttt{identity\_attack} or \texttt{insult} were categorised as Bias and Fairness risks. Threshold values for each toxicity category were set to systematically assign reports to risk types, ensuring that more severe risks were prioritised.

Using these mappings, we conducted simulations to compare the \textit{priority-based} and \textit{non-prioritised} supervision policies over a 12-month period, including a 3-month observation period to establish processing capacity. Note that the simulation runs beyond 12 months, continuing until the backlog of reports is emptied, reflecting a completion of all remaining tasks.

\begin{figure*}[htbp] 
    \centering 
    \includegraphics[width=0.6\textwidth]{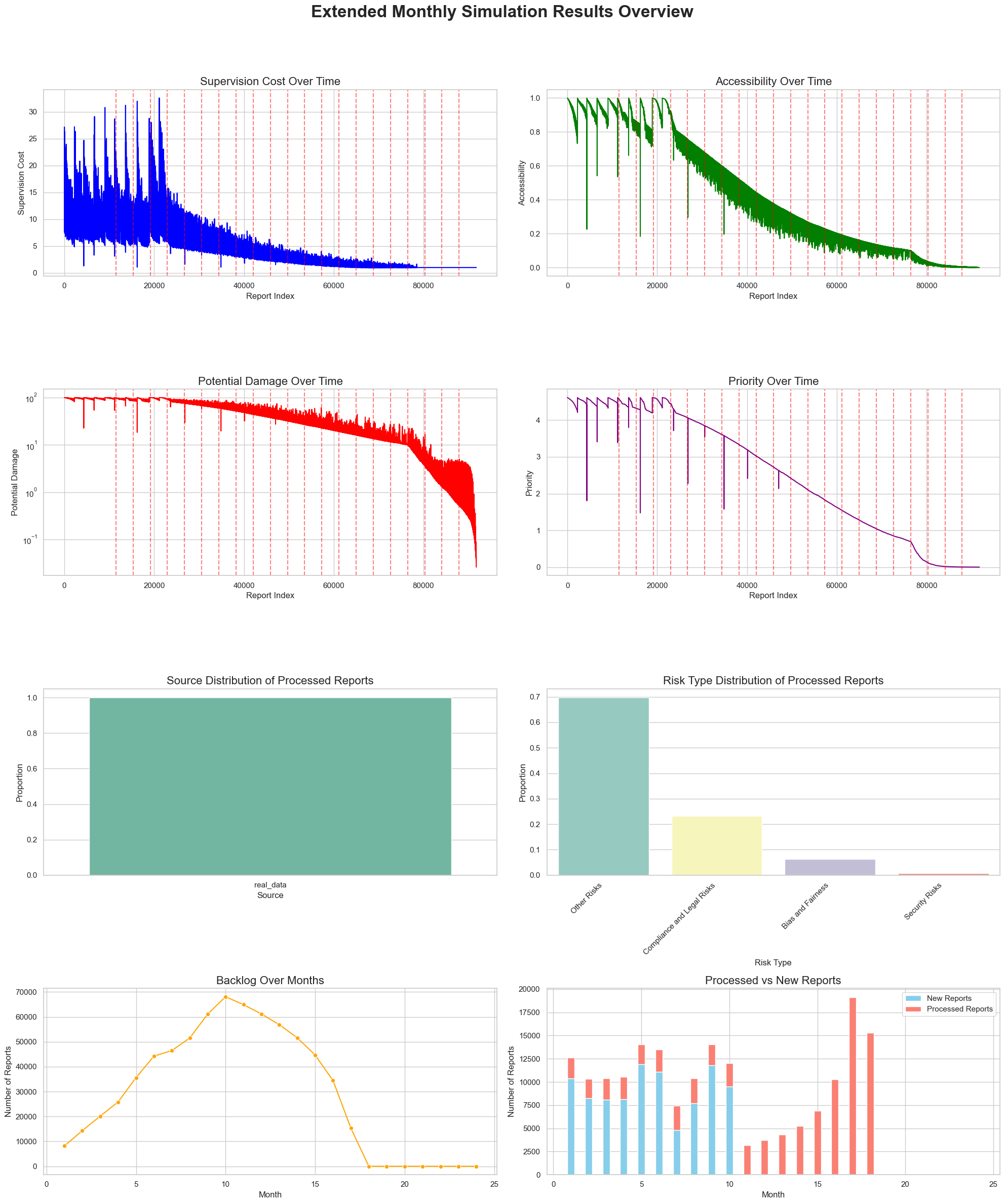} 
    \caption{Monthly Simulation Results Overview for the Prioritised Policy over the WildChat dataset. Simulation continues beyond the initial 12-month period to allow backlog processing to completion.} 
    \label{wildchat_prioritized} 
\end{figure*}

In the \textit{non-prioritised} policy (Figure~\ref{wildchat_non_prioritized}), supervision costs remain relatively stable over time, with frequent fluctuations. This indicates that the system processes reports in a first-come, first-served manner, addressing a wide variety of reports with differing supervision costs throughout the simulation. In contrast, the \textit{priority-based} policy (Figure~\ref{wildchat_prioritized}) shows a significant decline in supervision costs over time. This demonstrates that high-cost, high-impact reports are prioritised early in the simulation, and as those reports are addressed, the remaining ones are less resource-intensive, resulting in lower supervision costs toward the later stages.

For accessibility, the \textit{non-prioritised} policy (Figure~\ref{wildchat_non_prioritized}, top-right) maintains consistent scores throughout the simulation, reflecting that reports are processed in the order received, regardless of accessibility or detectability. In contrast, the \textit{priority-based} policy (Figure~\ref{wildchat_prioritized}, top-right) shows a decrease in accessibility over time, indicating that more accessible reports are addressed early, leaving harder-to-reach risks as the simulation progresses. Similar patterns are observed for potential damage and priority over time.

Regarding risk type distribution, the \textit{non-prioritised} policy (Figure~\ref{wildchat_non_prioritized}, bottom-right) processes a more balanced mix of risk types, while the \textit{priority-based} policy focuses on higher-impact categories such as Compliance and Legal Risks and Bias and Fairness, which are prioritised earlier due to their higher potential damage or priority scores.

\begin{figure*}[htbp]
    \centering
    \includegraphics[width=0.8\textwidth]{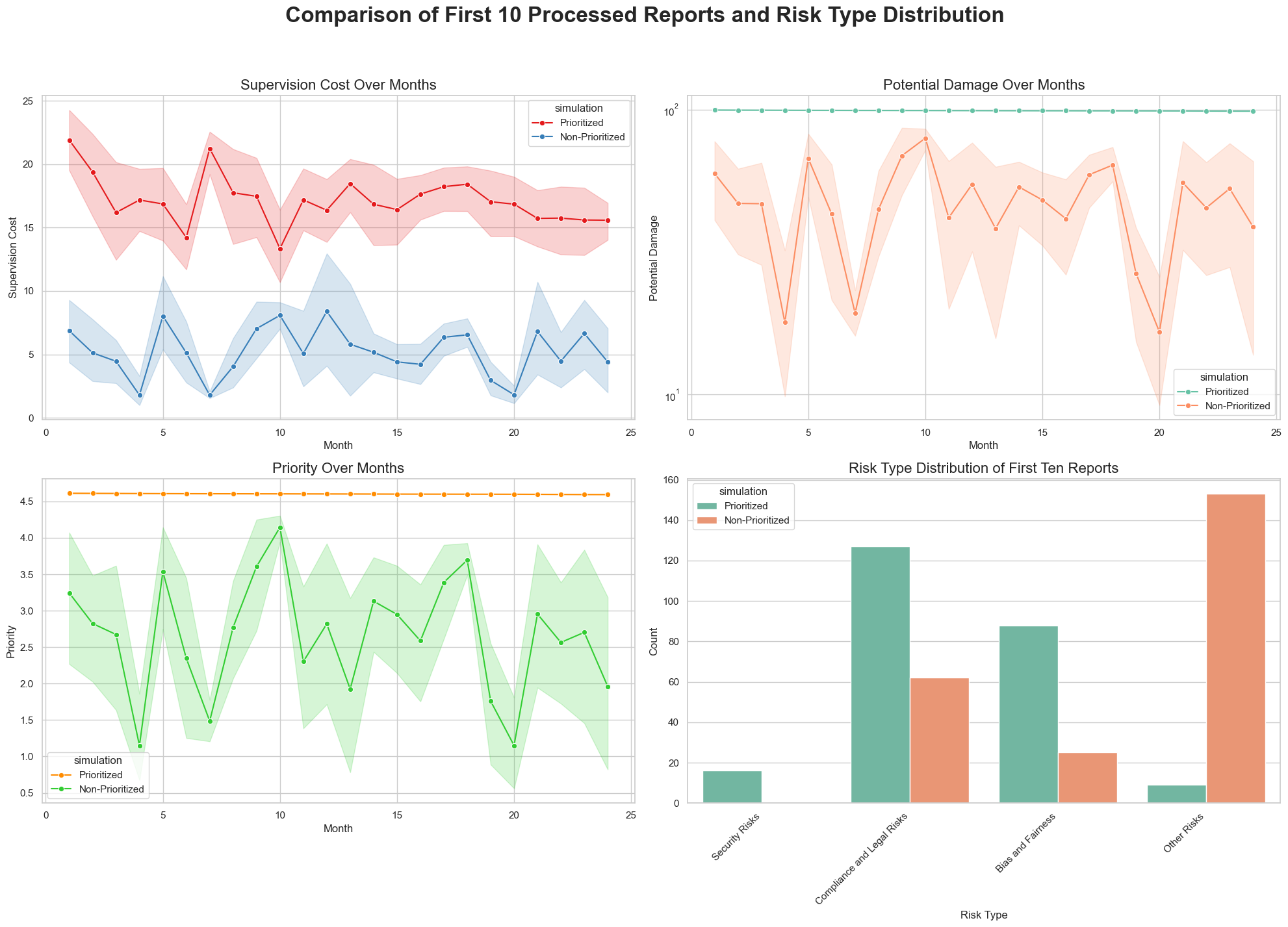}
    \caption{Comparison of the first 10 processed reports each month for both \textit{priority-based} and \textit{non-prioritised} approaches.}
    \label{fig:top_10}
\end{figure*}

As shown in Figure~\ref{fig:top_10}, the comparison of the first 10 processed reports each month for both \textit{priority-based} and \textit{non-prioritised} approaches reveals distinct differences. The \textit{priority-based} strategy incurs consistently higher supervision costs (top-left) and focuses on addressing high-potential-damage reports more consistently (top-right). In contrast, the \textit{non-prioritised} method fluctuates more in both supervision cost and potential damage. \textit{Priority-based} processing consistently targets high-priority reports (bottom-left), whereas the \textit{non-prioritised} approach handles reports with greater variability in priority. The risk type distribution (bottom-right) shows that while both methods handle similar types of risks, the \textit{priority-based} approach emphasises compliance and bias-related risks slightly more.

The \textit{priority-based} policy effectively manages high-priority risks, addressing the most urgent and impactful reports early in the simulation. In contrast, the \textit{non-prioritised} policy, while offering a balanced processing approach, struggles to address high-priority reports promptly.

To provide additional validation for our findings, we applied the same prioritisation framework to the ToxicChat dataset, a crowdsourced collection of reports annotated for toxicity and safety issues \cite{lin2023toxicchat}. The results, detailed in the Supplementary Information, mirrored those observed with the WildChat dataset: \textit{priority-based} processing focused on high-risk cases with narrower priority score distributions and timely handling of critical reports, while \textit{non-prioritised} processing exhibited broader distributions and delays in addressing high-priority content.

\section{DISCUSSION}

The findings of our simulation study underscore the profound impact that supervision policies can have on the management of risks associated with GPAI models. By systematically comparing \textit{non-prioritised}, \textit{random}, 
\textit{priority-based}, and \textit{diversity-prioritised} policies, we revealed how different strategies lead to markedly divergent outcomes in terms of risk mitigation efficiency and the coverage of diverse risk types.

Our results indicate that the \textit{priority-based} and \textit{diversity-prioritised} policies are more effective at addressing high-impact risks. Specifically, these policies processed reports with significantly higher potential damage and overall priority scores compared to the \textit{non-prioritised} and \textit{random} policies. For example, the average potential damage mitigated per report was substantially higher under the \textit{priority-based} approach, demonstrating its effectiveness in focusing resources on the most critical issues. However, this focus on high-priority reports introduces certain trade-offs. The \textit{priority-based} policy tended to favour reports from expert and crowdsourced sources, potentially overlooking valuable insights from the broader community.

The \textit{diversity-prioritised} policy, while offering a more balanced coverage across different risk types and sources, still prioritises high-impact risks, leaving room for overlooked community-driven reports that often cover emergent behaviours or user experience issues. The influence of feedback loops between incentives and deterrence further accentuates these trends. As more attention is given to expert-driven reports, the incentives for experts to continue reporting increase, reinforcing their influence. Meanwhile, community and crowdsourcing sources, receiving less attention, experience reduced incentives, leading to a decline in their reporting. This dynamic highlights the challenge of balancing short-term prioritisation with long-term risk diversity.

To validate these findings, we applied the simulation framework to the WildChat dataset, which comprises over one million conversations with ChatGPT, including over 150K conversations flagged as risky. This real-world data provided empirical support for our simulations, reinforcing the efficiency of the \textit{priority-based} policy in addressing high-impact risks, particularly those related to compliance and legal issues or bias and fairness. The validation showed similar patterns to those observed in the simulations, where high-priority reports received timely attention, while lower-priority issues were more likely to be delayed or overlooked.

From an ethical and societal perspective, our study highlights the importance of inclusive and diverse risk management practices by AI supervisory entities. Policies that overly prioritise expert reports may inadvertently marginalise community voices, leading to a narrower understanding of the risk landscape. Given that community members often experience and report issues directly affecting end-users, their input is crucial for a comprehensive risk mitigation strategy. AI supervisory entities should consider incorporating mechanisms that value diverse perspectives to avoid blind spots in their oversight.

It is important to note the limitations of our study. While our simulations and real-world data validation provide valuable insights, they remain simplified models of highly complex processes. Future research should gather more empirical data from AI supervisory entities to validate and refine the simulation parameters further. This could include detailed case studies and longitudinal studies tracking the long-term outcomes of various supervision strategies. Additionally, our work assumes that AI supervisory entities will actively engage with and prioritise AI risk management efforts as proposed in this framework. However, it is possible that such supervision could evolve in ways that diverge from these assumptions, potentially leading to a more self-sufficient or automated future where the need for external oversight might be reduced.

While our analysis highlights the potential benefits and trade-offs of various risk supervision policies, implementing these strategies within actual supervisory bodies presents multiple challenges. Determining priority scores, for instance, requires operationalising concepts such as accessibility and potential damage, which may demand specialised technical expertise, extensive training data, and periodic recalibration as new types of risks emerge. Human resource considerations, including the need for dedicated analysts, risk assessors, and legal advisors, can significantly increase operational costs. Strict requirements for explainability, auditing, and redress can influence the feasibility of certain prioritisation methods and may necessitate advanced documentation or transparency tools. Ethical and legal constraints, including data protection, privacy considerations and fairness standards, further limit the scope of feasible implementations. Ensuring that the chosen policy meets both the technical and normative criteria of evolving regulatory landscapes—and can be explained to stakeholders—will be essential for bridging the gap between theoretical policy design and sustainable, responsible real-world governance.

Despite these limitations, our study contributes to illuminating the critical yet understudied aspect of GPAI models' risk supervision policies, opening up a crucial area of investigation: their downstream effects on incentives, risk deterrence, and the broader risk landscape.

\section*{Code and Data Availability}

The simulation code for this study is available in the GitHub repository:

\url{https://github.com/manuelcebrianramos/LLM_supervision_policies}

This repository contains the following:
\begin{itemize}
    \item The Jupyter Notebook (\texttt{Simulation\_Code/\\PrioritizationLLMsupervision.ipynb}) implementing the Report class and associated simulation functions.
    \item All simulation data organised into batch folders in \texttt{Simulation\_Runs/}.
\end{itemize}

The folder \texttt{Simulation\_Runs/} contains CSV files for each simulation type, named according to the following patterns:
\begin{itemize}
    \item \small \texttt{simulation.YYYYMMDD\_HHMMSS.csv} for general simulations,
    \item \small \texttt{non-prioritized\_simulation.XX\_YYYYMMDD\_HHMMSS.csv} for non-prioritised simulations,
    \item \small \texttt{random\_fairness\_simulation.XX\_YYYYMMDD\_HHMMSS.csv} for random fairness simulations,
    \item \small \texttt{prioritized\_simulation.XX\_YYYYMMDD\_HHMMSS.csv} for priority-based simulations,
    \item {\footnotesize \texttt{diversity\_prioritized\_simulation.XX\_YYYYMMDD\_HHMMSS.csv}} for diversity-prioritised simulations.
\end{itemize}

Each file contains detailed results of an individual simulation run. Simulation results are further organised into batch folders, such as \texttt{First\_batch} and \texttt{Second\_batch}, for ease of access and analysis based on different experimental waves. The results discussed in this paper are derived from the \texttt{Second\_batch}, which contains the most up-to-date experimental data.

In addition to these simulations, we integrated datasets that further enrich our analysis of incident trends. These datasets include:

\begin{itemize}
    \item ToxicChat Dataset, a crowdsourced dataset from the Vicuna online demo, containing 10,165 annotated reports for toxicity, jailbreaking, and OpenAI moderation scores, along with human annotations. This dataset served as the primary foundation for our prioritisation analysis. It is available at \url{https://huggingface.co/datasets/lmsys/toxic-chat}.
    \item OECD AI Incidents Monitor (AIM), utilised for projecting long-term trends in AI incidents and hazards. It can be accessed at \url{https://oecd.ai/en/incidents}.
    \item Jailbreak prompt data from Shen et al. (2023) \cite{shen2023jailbreak}, which tracks jailbreak attempts across online communities, illustrating trends before and after a major regulatory intervention in June 2023. This dataset is available at \url{https://github.com/verazuo/jailbreak_llms}.
\end{itemize}

The WildChat dataset used for validating our simulations is accessible via Hugging Face at:

\url{https://huggingface.co/datasets/allenai/WildChat-1M-Full}

This dataset contains one million conversations between users and ChatGPT (versions 3.5 and 4), spanning multiple languages and interaction turns, with a subset flagged as containing toxic or risky content. Combined with the ToxicChat dataset, this dataset helped ensure robustness in our prioritisation simulations and moderation strategies.

\bibliographystyle{unsrt}
\bibliography{references}

\section*{Disclaimer}
The views expressed in this article are solely those of the authors and should not, under any circumstances, be regarded as representing the official positions of the European Commission.

\section*{Acknowledgments}
This work was primarily funded by the HUMAINT project, supported by the Directorate-General Joint Research Centre of the European Commission. 

During the preparation of this work the author(s) used Google Gemini, Anthropic Claude, and OpenAI ChatGPT in order to proof-read the manuscript’s language and assist with small code snippets/debugging in the simulation notebook. After using these tools, the authors reviewed and edited the content as needed and take full responsibility for the content of the publication.

\section*{Supplementary Information}

\subsection*{Understanding Reporting Sources of GPAI Model Systemic Risks: Community Reporting, Crowdsourcing, and Expert Groups}

In this subsection we examine three significant approaches for documenting GPAI model systemic risks: community reporting, crowdsourcing, and expert groups.

To assess different approaches to addressing GPAI model-induced harm, we focused on three core methods. First, we conducted a thorough examination of existing studies and reports on community reporting, with a particular focus on online communities dedicated to tracking and documenting GPAI model issues. This included analysis of platforms such as Reddit, Discord, and specialized websites like AIPRM and FlowGPT. Second, we analysed the outcomes of relevant open challenges, including hackathons, crowdsourced open platforms, and corporate collaborative research programs. This involved studying initiatives like the Chatbot Arena, SafeBench, and OpenAI's Preparedness Challenge. Third, we evaluated insights from academic expert groups that bring together AI safety scientists to document GPAI model vulnerabilities in a structured manner. This included reviewing work from groups such as those behind the WMDP Benchmark and the Center for AI Safety. Our research methodology also incorporated analysis of available datasets and relevant scholarly work related to these approaches. 

By examining these strategies, this subsection aims to provide an overview of effective methods for understanding and mitigating the systemic risks posed by GPAI model technology. In what follows, we delve into these three distinct citizen science strategies designed to study and mitigate harm caused by GPAI models.\\ 

\subsubsection{Community-Driven Reporting of GPAI Model Harm}

The community-driven approach leverages the collective intelligence and vigilance of online communities dedicated to tracking and documenting GPAI model vulnerabilities. This method exemplifies the power of distributed, bottom-up monitoring in the rapidly evolving landscape of AI safety. The work \cite{shen2023jailbreak} serves as a cornerstone in understanding the efficacy of community-driven reporting. Their comprehensive study, analysing over 600 jailbreak prompts from platforms like Reddit and Discord, showcases the potential of this approach. Jailbreak prompts are specifically crafted inputs designed to bypass a GPAI model's ethical safeguards, causing it to generate content that would normally be restricted. By combining advanced NLP techniques with meticulous manual curation, the study \cite{shen2023jailbreak} unveiled a spectrum of strategies used to circumvent GPAI model safety mechanisms, ranging from straightforward exploitation attempts to sophisticated adversarial attacks designed to undermine ethical guidelines.

Platforms such as Reddit, Discord, and X (formerly Twitter) provide an ideal ecosystem for community-centric methodologies. Their decentralized nature fosters unconstrained user engagement and the formation of highly specialized communities. For instance, the \texttt{r/ChatGPT} subreddit boasts over 2.3 million users, while more specialized communities like \texttt{r/ChatGPTPromptGenius} and \texttt{r/ChatGPTJailbreak} have 97,500 and 13,500 users respectively. These digital spaces act as incubators for rapid dissemination and iterative refinement of tactics aimed at probing GPAI model capabilities and limitations. The diversity of perspectives and technical expertise within these communities catalyses innovation. This dynamic creates a self-perpetuating ecosystem that drives continuous vulnerability identification and, paradoxically, fuels the development of more robust countermeasures.

Beyond general social media platforms, specialized websites like AIPRM, FlowGPT, and JailbreakChat have emerged as dedicated hubs for GPAI model prompt engineering and testing. AIPRM, for example, is a ChatGPT extension with approximately 1 million users, offering a curated collection of prompts, including some designed for jailbreaking. These platforms serve a dual purpose: they provide a concentrated environment for users to develop and refine prompts that probe GPAI model limitations, and they offer researchers a rich source of data on real-world GPAI model interactions and potential vulnerabilities. The open-source nature of many community-driven initiatives further amplifies their impact. Datasets like AwesomeChatGPTPrompts, with 163 documented prompts, and the OCR Twitter/Reddit Dataset, containing 50 prompts, provide researchers with extensive corpora of user-generated content, reflecting authentic usage patterns and challenges faced by GPAI models in production environments.

A critical observation by \cite{shen2023jailbreak} is the migration of sensitive prompts and exploits from public forums to more secluded online spaces. This trend highlights the dynamic nature of the threat landscape and underscores the need for adaptive monitoring strategies. Community-driven approaches are uniquely positioned to track these shifts, providing early warnings and insights that might elude more formal, structured monitoring systems.

The organic nature of online AI safety communities presents both opportunities and challenges for researchers and policymakers. Community members are often driven by genuine interest and concern, leading to high engagement and novel insights. These communities bring together individuals from various backgrounds, fostering interdisciplinary approaches to AI safety. The fluid nature of online communities allows for quick responses to new AI developments and emerging threats. However, the spontaneous and decentralised nature of these communities makes it difficult to replicate their dynamics in controlled environments or through incentive-driven structures. Additionally, the dual-use nature of jailbreaking research raises ethical questions about the responsible disclosure and discussion of vulnerabilities.

Community reporting offers several distinct advantages. It provides real-time data collection, enabling immediate insights into new vulnerabilities as they emerge. The diverse perspectives within these communities enhance the depth and breadth of analysis, capturing a more comprehensive picture of GPAI model impacts. Utilising existing community structures makes this approach cost-effective, scalable, and sustainable. The agility of these communities allows for quick adaptation to new AI developments and shifting threat landscapes. Furthermore, the unstructured nature of community interactions can lead to serendipitous discoveries and valuable insights that might be overlooked in more formal research settings.

However, community reporting also faces significant challenges. The open nature of these platforms can lead to inconsistencies in data quality, necessitating rigorous validation processes. There are notable geographical biases, with the dominance of certain regions in tech communities potentially skewing perspectives and overlooking issues relevant to underrepresented areas. Standardisation is a major hurdle, as the variability in reporting formats complicates data aggregation and analysis. Community-driven research may sometimes operate in ethically or legally ambiguous territories, raising concerns about responsible practices. Additionally, without proper verification mechanisms, there's a risk of false or misleading information about GPAI model vulnerabilities spreading within these communities.

The real-time, diverse, and adaptive nature of community reporting makes it an invaluable component of a holistic AI safety ecosystem. 

\begin{table*}[ht]
\centering
\caption{Identified Community-Driven Reporting Platforms}
\label{tab:community_reporting}
\resizebox{\textwidth}{!}{%
\begin{tabular}{|>{\raggedright\arraybackslash}p{4cm}|>{\raggedright\arraybackslash}p{2.5cm}|>{\raggedright\arraybackslash}p{2.5cm}|>{\raggedright\arraybackslash}p{5cm}|>{\raggedright\arraybackslash}p{3.5cm}|}
\hline
\textbf{Community Name} & \textbf{Platform} & \textbf{Type} & \textbf{Focus} & \textbf{Approx. Size (If Available)} \\
\hline
r/ChatGPT & Reddit & Subreddit & General ChatGPT Discussion & 2.3M users \\
\hline
r/ChatGPTPromptGenius & Reddit & Subreddit & Prompt Sharing (with jailbreak emphasis) & 97.5K users \\
\hline
r/ChatGPTJailbreak & Reddit & Subreddit & Jailbreak Prompts & 13.5K users \\
\hline
ChatGPT & Discord & Server & & \\
\hline
ChatGPT Prompt Engineering & Discord & Server & & \\
\hline
Spreadsheet Warriors & Discord & Server & & \\
\hline
AI Prompt Sharing & Discord & Server & & \\
\hline
GPAI model Promptwriting & Discord & Server & & \\
\hline
BreakGPT & Discord & Server & & \\
\hline
AIPRM & Website & ChatGPT Extension & Curated Prompts (some jailbreak) & 1M users \\
\hline
FlowGPT & Website & Community Prompt Sharing & Prompts with User Tags (some jailbreak) & \\
\hline
JailbreakChat & Website & Dedicated Jailbreak Prompts & & \\
\hline
{\small AwesomeChatGPTPrompts} & Open-Source & Dataset & General Prompts & 163 prompts \\
\hline
OCR Twitter/Reddit Dataset & Open-Source & Dataset & General Prompts (some may be jailbreak) & 50 prompts \\
\hline
\end{tabular}%
}
\end{table*}

\subsubsection{Crowdsourced Reporting of GPAI Model Harm}

Crowdsourcing harnesses the collective power of large, diverse groups to identify and mitigate harms caused by GPAI models. This approach engages the public through incentive-based challenges and competitions, mobilising a vast array of participants who bring unique perspectives and expertise to the task. While distinct from community-driven reporting, crowdsourcing complements it by providing structured frameworks for problem-solving and innovation.

The DARPA Network Challenge exemplifies the potential of crowdsourcing to rapidly mobilize collective intelligence. Participants were tasked with locating red weather balloons scattered across the United States, with monetary rewards spurring swift and effective collaboration among a wide network of individuals and teams. This challenge demonstrated the power of incentives to efficiently harness collective resources \cite{pickard2011time,rutherford2020impossible}. 

Crowdsourcing platforms foster innovation by encouraging creative problem-solving. The Chatbot Arena, for instance, employs a crowdsourced, randomized battle platform where users evaluate GPAI models through direct interaction. This platform utilizes the Elo rating system, commonly used in competitive games like chess, to rank GPAI models based on user preferences. Such innovative evaluation techniques provide valuable insights into model performance and user satisfaction \cite{chiang2023vicuna}.

Engaging a broad participant base brings diverse perspectives and expertise to the forefront. Platforms like Kaggle host competitions that attract data scientists and AI enthusiasts worldwide, fostering a community-driven approach to problem-solving. Kaggle competitions often involve tasks such as identifying biases in AI models or developing new algorithms to improve model performance. The diversity of participants enhances the quality and breadth of solutions generated through these competitions.

Several case studies illustrate the effectiveness of crowdsourcing in GPAI model evaluation and safety. The Chatbot Arena evaluates GPAI models using crowdsourced user interactions, offering dynamic and continuous assessment of GPAI model performance in real-world scenarios. SafeBench represents a structured effort focused on AI safety evaluation, engaging participants in identifying and addressing safety concerns related to GPAI models. OpenAI's Preparedness Challenge is designed to engage the community in proactive safety measures, emphasizing collaboration and ethical considerations. Kaggle's `GPAI model Safeguards and Guardrails'' project further demonstrates how competitive environments can foster innovative solutions to AI safety challenges.

Citizen science offers unique advantages in tackling the challenges of identifying and mitigating GPAI model harm. It facilitates analysis of vast data volumes, provides diverse viewpoints for detecting biases, privacy concerns, and misinformation, and enables real-world GPAI model testing across various scenarios and languages -- areas often overlooked in traditional research. Moreover, the rapid proliferation and diversification of GPAI models across developers, vendors, and iterations emphasize the complexity of the AI landscape. This underscores the value of citizen science in providing a ground-level perspective for observing, analysing, and mitigating the multifaceted impacts of these technologies.

\begin{table*}[ht]
\centering
\caption{Identified Crowdsourced Reporting Case Studies}
\label{tab:crowdsourcing_case_studies}
\begin{tabular}{|p{3cm}|p{3cm}|p{5cm}|p{5cm}|}
\hline
\textbf{Case Study} & \textbf{Platform} & \textbf{Methodology} & \textbf{Strengths and Limitations} \\
\hline
\textbf{Chatbot Arena} & LMSYS Org & Benchmarking GPAI models in real-world scenarios using crowdsourced user interactions and Elo rating system & \textbf{Strengths:} Real-time feedback, user-driven evaluation, scalable.\\
& & & \textbf{Limitations:} Potential for bias in user preferences, requires active engagement. \\
\hline
\textbf{SafeBench} & MLSafety.org & Evaluating AI safety through structured tasks and collaborative environments & \textbf{Strengths:} Innovation in safety solutions, extensive testing and stress-testing.\\
& & & \textbf{Limitations:} Requires careful task design, dependent on participant expertise. \\
\hline
\textbf{OpenAI's Preparedness Challenge} & OpenAI & Developing safeguards and identifying vulnerabilities through incentivized exploration of risks and vulnerabilities & \textbf{Strengths:} Promotes collaboration, addresses evolving risks, emphasizes ethical considerations.\\
& & & \textbf{Limitations:} Long-term impact uncertain, requires continuous monitoring. \\
\hline
\textbf{Kaggle Competitions} & Kaggle & Competitive data science tasks fostering collaboration and innovation in GPAI safety & \textbf{Strengths:} Diverse contributions, fosters innovation, extensive knowledge sharing.\\
& & & \textbf{Limitations:} Competitive nature may limit collaboration, dependent on data quality. \\
\hline
\end{tabular}
\end{table*}

\subsubsection{Expert Group Reporting of GPAI Model Harm}

Expert groups play a vital role in identifying, analysing, and reporting potential harms caused by GPAI models. These groups, composed of specialists from various fields including AI ethics, computer science, linguistics, and domain-specific areas like biosecurity and cybersecurity, offer unique insights that complement other approaches to GPAI model safety evaluation. The depth of their analyses and the rigour of their methodologies make expert group reporting an essential component in understanding and mitigating GPAI model-related risks.

The MLCommons AI Safety Benchmark v0.5 \citep{vidgen2024introducing} exemplifies the comprehensive approach that expert groups can take in evaluating GPAI model safety. Developed by a consortium of industry and academic researchers, this benchmark introduces a new taxonomy of 13 hazard categories, providing a structured framework for assessing GPAI model risks. The benchmark's focus on specific use cases and personas demonstrates how expert groups can create targeted, yet broadly applicable, evaluations of GPAI model safety. By developing a grading system that is open, explainable, and adjustable for various use cases, the MLCommons team showcases how expert groups can create flexible tools for assessing GPAI model safety across different contexts and applications.

Another significant contribution to expert group reporting is the WMDP Benchmark \citep{li2024wmdp}, which focuses on measuring GPAI model capabilities related to weapons of mass destruction. This benchmark, created by a focus group of experts, provides a quantitative framework for assessing GPAI models' potential to facilitate the development of dangerous biological, cyber, and chemical weapons. It demonstrates how expert groups can move beyond anecdotal risk assessments towards rigorous, quantitative evaluations of potential misuse. The WMDP Benchmark's approach to creating proxy measurements for hazardous knowledge while adhering to strict ethical guidelines illustrates the unique ability of expert groups to navigate complex ethical considerations in AI safety research.

The studies \cite{gopal2023pandemic,gopal2023releasing} offer a different approach to expert group reporting, focusing on a specific high-risk scenario. Their work, involving a simulated scenario where participants acted as bioterrorists using a GPAI model to obtain information about the 1918 influenza virus, revealed nuanced insights into how GPAI models could be manipulated for nefarious purposes. This study highlights the ability of expert groups to design sophisticated test scenarios that probe the boundaries of GPAI model capabilities and safety measures. By demonstrating how GPAI models could lower the barrier of entry for individuals seeking to cause harm, even if they lack specialized knowledge. The study \cite{gopal2023pandemic,gopal2023releasing} underscores the importance of considering broader contextual factors in GPAI model risk assessment.

These examples of expert group reporting share several key strengths. They all employ rigorous methodologies, ensuring the reliability and reproducibility of their findings. The MLCommons and WMDP benchmarks showcase the ability of expert groups to create comprehensive frameworks for evaluating GPAI model safety, while the study \cite{gopal2023pandemic,gopal2023releasing}  demonstrates how focused scenarios can reveal subtle vulnerabilities. All three examples navigate complex ethical considerations, balancing the need for thorough evaluation with responsible disclosure practices.

In conclusion, expert group reporting of GPAI model harm, as exemplified by the MLCommons AI Safety Benchmark, the WMDP Benchmark, and the studies \cite{gopal2023pandemic,gopal2023releasing}, offer valuable, in-depth analyses of GPAI model-related risks. These efforts provide crucial insights for developing effective safety measures and policies, forming a cornerstone of comprehensive GPAI model safety evaluation strategies. As GPAI model technologies continue to advance, the role of expert groups in safety assessment and harm reporting will likely become increasingly critical, providing a foundation for responsible AI development and deployment.

\begin{table*}[ht]
\centering
\caption{Identified Expert Group Reporting Case Studies}
\label{tab:expert_group_case_studies}
\begin{tabular}{|>{\raggedright\arraybackslash}p{3cm}|>{\raggedright\arraybackslash}p{4cm}|>{\raggedright\arraybackslash}p{3cm}|>{\raggedright\arraybackslash}p{4cm}|>{\raggedright\arraybackslash}p{4cm}|}
\hline
\textbf{Case Study} & \textbf{Platform} & \textbf{Methodology} & \textbf{Strengths} & \textbf{Limitations} \\
\hline
MLCommons AI Safety Benchmark v0.5 \citep{vidgen2024introducing} & MLCommons & Comprehensive taxonomy of 13 hazard categories; Multiple-choice questions across various scenarios & Broad coverage of potential harms; Structured, reproducible evaluation framework; Adjustable grading system & Resource-intensive to develop and update; May not capture all real-world complexities \\
\hline
WMDP Benchmark \citep{li2024wmdp} & Consortium of academics and technical consultants & Quantitative framework focused on weapons of mass destruction knowledge; Multiple-choice questions & Rigorous evaluation of specific high-risk domains; Balances comprehensive assessment with ethical considerations & Narrow focus on specific threat vectors; Resource-intensive to develop and administer \\
\hline
Gopal et al. Study \citep{gopal2023pandemic} & Academic research group & Simulated scenario of bioterrorists using GPAI model; Qualitative analysis of participant behaviour & Deep insights into specific high-risk scenarios; Reveals nuanced manipulation of GPAI models & Limited scope; May not generalize to all types of GPAI model misuse \\
\hline
\end{tabular}
\end{table*}

\subsubsection{Summary}

The rapid advancement and widespread deployment of GPAI models demand a comprehensive and multifaceted approach to identifying, reporting, and mitigating their potential harms. This subsection has explored three primary strategies: community-driven reporting, crowdsourcing, and expert group evaluations, each bringing distinct strengths to the complex landscape of GPAI model safety. Community-driven reporting excels in providing broad coverage and real-time responsiveness, leveraging diverse perspectives to track emerging vulnerabilities. Crowdsourcing fosters structured and scalable problem-solving, mobilizing collective intelligence to address specific challenges with innovative solutions. Expert groups offer unparalleled depth and rigour, delivering authoritative analyses and recommendations for tackling the most complex and technical issues. The complementary nature of these approaches underscores the need for an integrated strategy to ensure robust and adaptive GPAI safety mechanisms.

\subsection*{Priority Score Illustration}

\begin{figure}[H]
    \centering
    \includegraphics[width=0.5\textwidth]{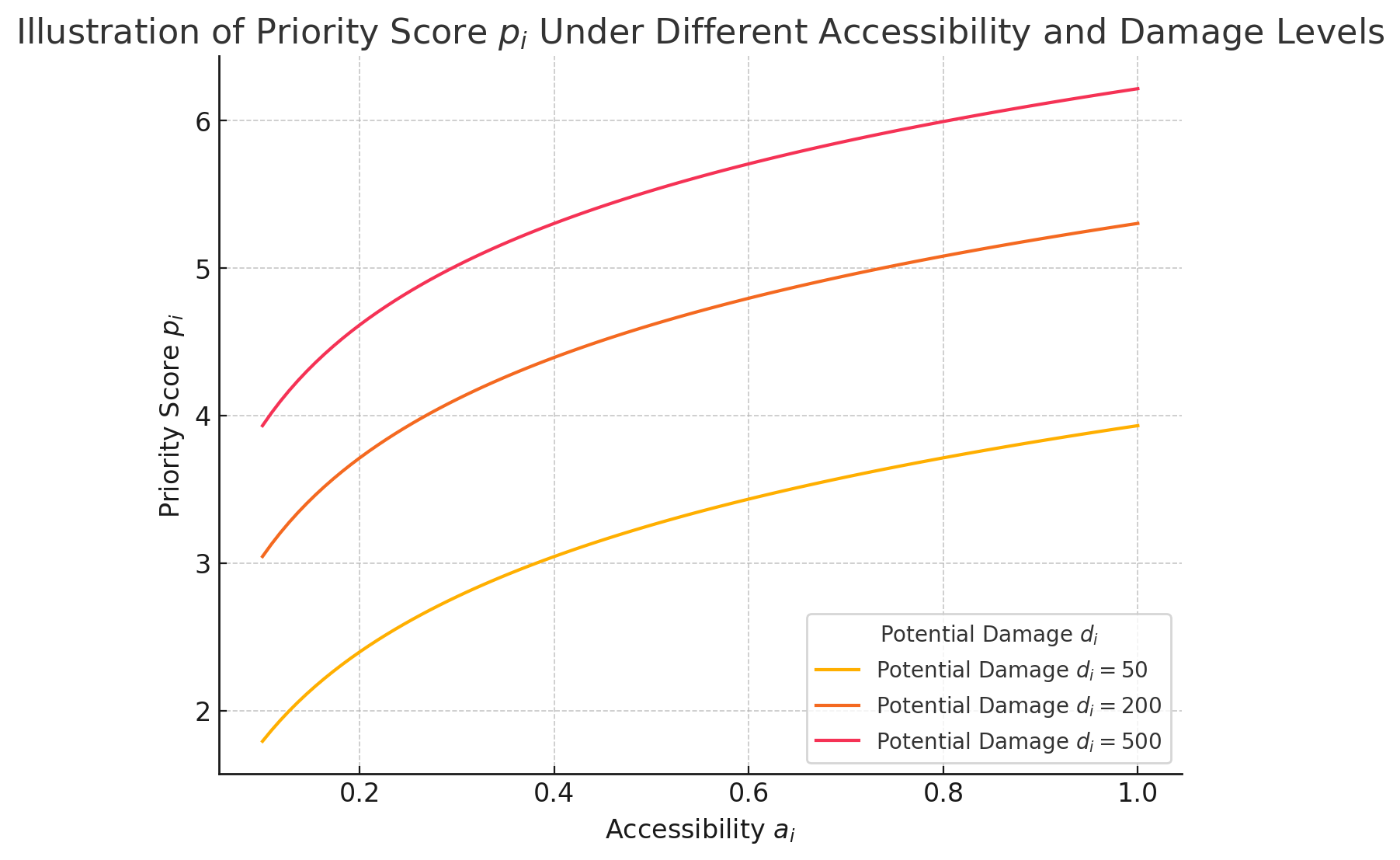}
    \caption{Illustration of priority score \( p_i \) under different combinations of accessibility \( a_i \) and potential damage \( d_i \), highlighting the logarithmic scaling effect.}
    \label{fig:priority_illustration}
\end{figure}

Figure \ref{fig:priority_illustration} illustrates how the priority score \( p_i = \log(1 + a_i \cdot d_i) \) responds to varying values of accessibility \( a_i \) and potential damage \( d_i \). This visualisation demonstrates the logarithmic scaling, showing that while high accessibility and damage increase priority, extreme values do not disproportionately dominate the score.

\subsection*{Report Generation Distributions}

\begin{figure}[H]
    \centering
    \includegraphics[width=0.5\textwidth]{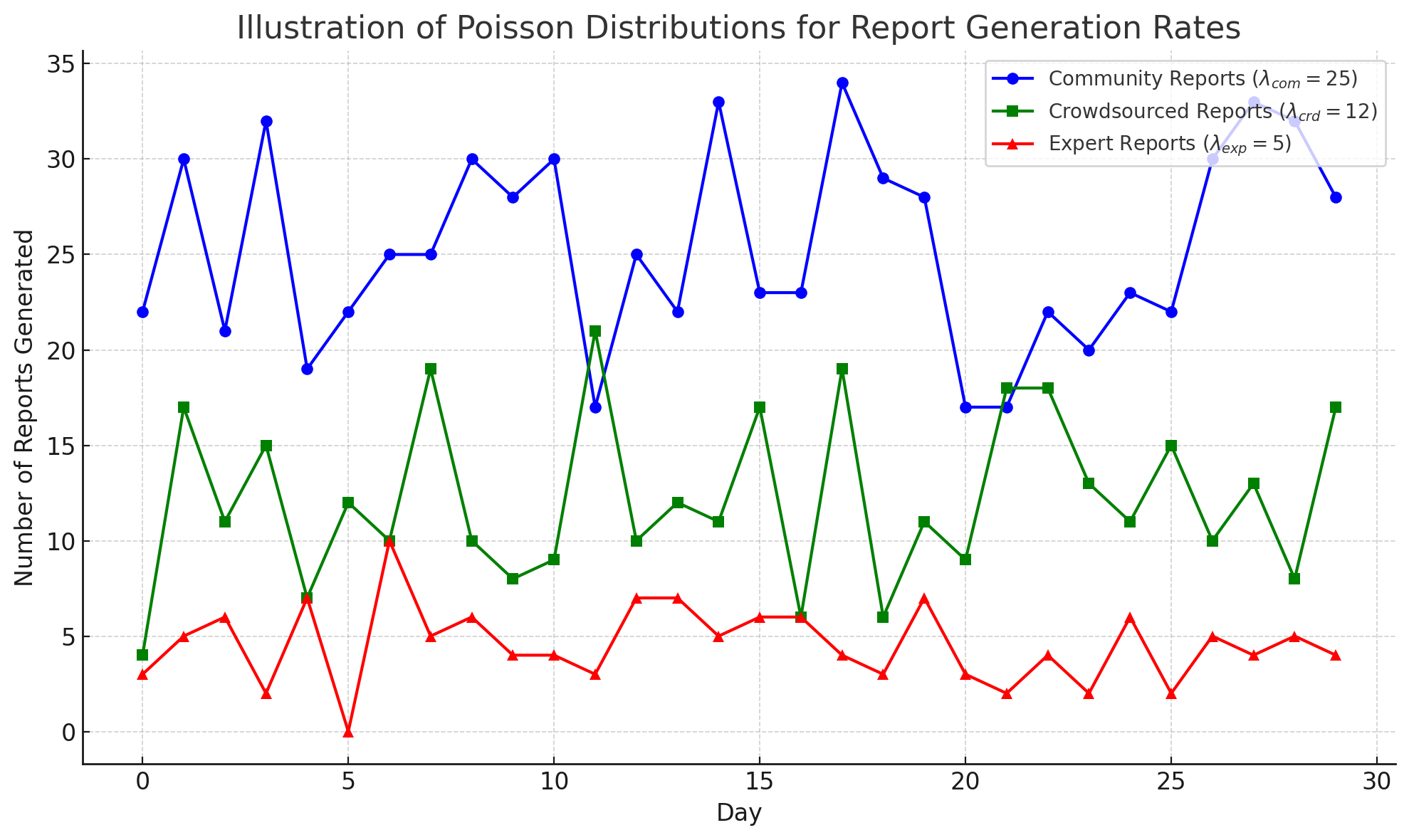}
    \caption{Illustration of Poisson distributions for report generation rates: community reports (\(\lambda_{\text{com}} = 25\)), crowdsourced reports (\(\lambda_{\text{crd}} = 12\)), and expert reports (\(\lambda_{\text{exp}} = 5\)).}
    \label{fig:report_generation_distribution}
\end{figure}

Figure \ref{fig:report_generation_distribution} provides a visual representation of the Poisson distributions used for each source type’s report generation rate. These distributions reflect the varying report frequencies among community, crowdsourced, and expert sources, emphasizing the higher volume from community sources and the more selective nature of expert reports.

\subsection*{Parameter Distributions}

\begin{figure}[H]
    \centering
    \includegraphics[width=0.4\textwidth]{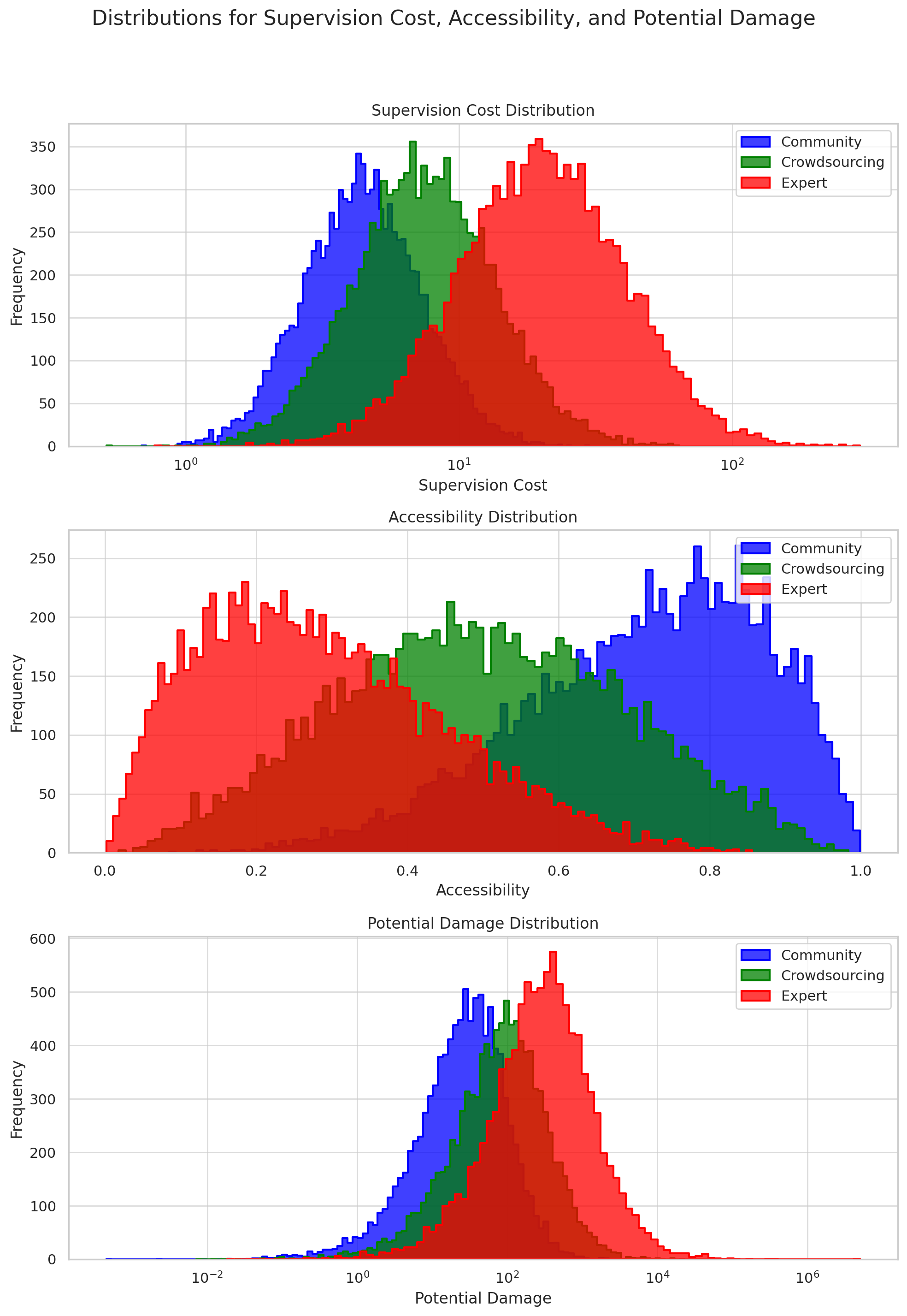}
       \caption{Illustration of the source\textendash specific statistical distributions used by the simulator. Supervision cost $s_i$ follows log\textendash normal distributions calibrated to reflect increasing complexity: community $(\mu_{\text{com}} = 1.5,\,\sigma_{\text{com}} = 0.5)$, crowdsourced $(\mu_{\text{crd}} = 2.0,\,\sigma_{\text{crd}} = 0.6)$, and expert $(\mu_{\text{exp}} = 3.0,\,\sigma_{\text{exp}} = 0.7)$. 
Accessibility $a_i$ is modelled with beta distributions capturing varying ease of exploitation: community $(\alpha_{\text{com}} = 5,\,\beta_{\text{com}} = 2)$, crowdsourced $(\alpha_{\text{crd}} = 3,\,\beta_{\text{crd}} = 3)$, and expert $(\alpha_{\text{exp}} = 2,\,\beta_{\text{exp}} = 5)$. 
Potential damage $d_i$ follows Pareto distributions with scale parameters tuned per source: community $(k_{\text{com}} = 100)$, crowdsourced $(k_{\text{crd}} = 150)$, and expert $(k_{\text{exp}} = 250)$. 
These distributions jointly define each incident's stochastic profile in the subsequent simulation experiments.}
    \label{fig:parameter_illustration}
\end{figure}

Figure \ref{fig:parameter_illustration} provides a visual representation of the distributions used for each parameter (supervision cost, accessibility, and potential damage) across community, crowdsourced, and expert sources. The distinct colours and logarithmic scales applied to supervision cost and potential damage enhance the clarity of differences between each report type, reflecting their unique characteristics.

\begin{table}[H]                                  
\centering
\caption{Maximum-likelihood parameter estimates used in the simulator.
Log-normal parameters $(\mu,\sigma)$ model supervision cost $s_i$; Beta
parameters $(\alpha,\beta)$ model accessibility $a_i$; and the Pareto
scale $k$ models potential damage $d_i$.}
\label{tab:fitted_params}
\footnotesize            
\setlength{\tabcolsep}{4pt}
\begin{tabular}{lccccc}
\toprule
\textbf{Source} & $\boldsymbol{\mu}$ & $\boldsymbol{\sigma}$ &
$\boldsymbol{\alpha}$ & $\boldsymbol{\beta}$ & $\boldsymbol{k}$ \\
\midrule
Community     & 1.5 & 0.5 & 5 & 2 & 100 \\
Crowdsourced  & 2.0 & 0.6 & 3 & 3 & 150 \\
Expert        & 3.0 & 0.7 & 2 & 5 & 250 \\
\bottomrule
\end{tabular}
\end{table}

Table~\ref{tab:fitted_params} lists the maximum-likelihood parameters
used in the simulator.

\section*{Priority Distribution Analysis for the Crowdsourced ToxicChat Dataset}

\begin{figure}[h!]
    \centering
    \includegraphics[width=0.5\textwidth]{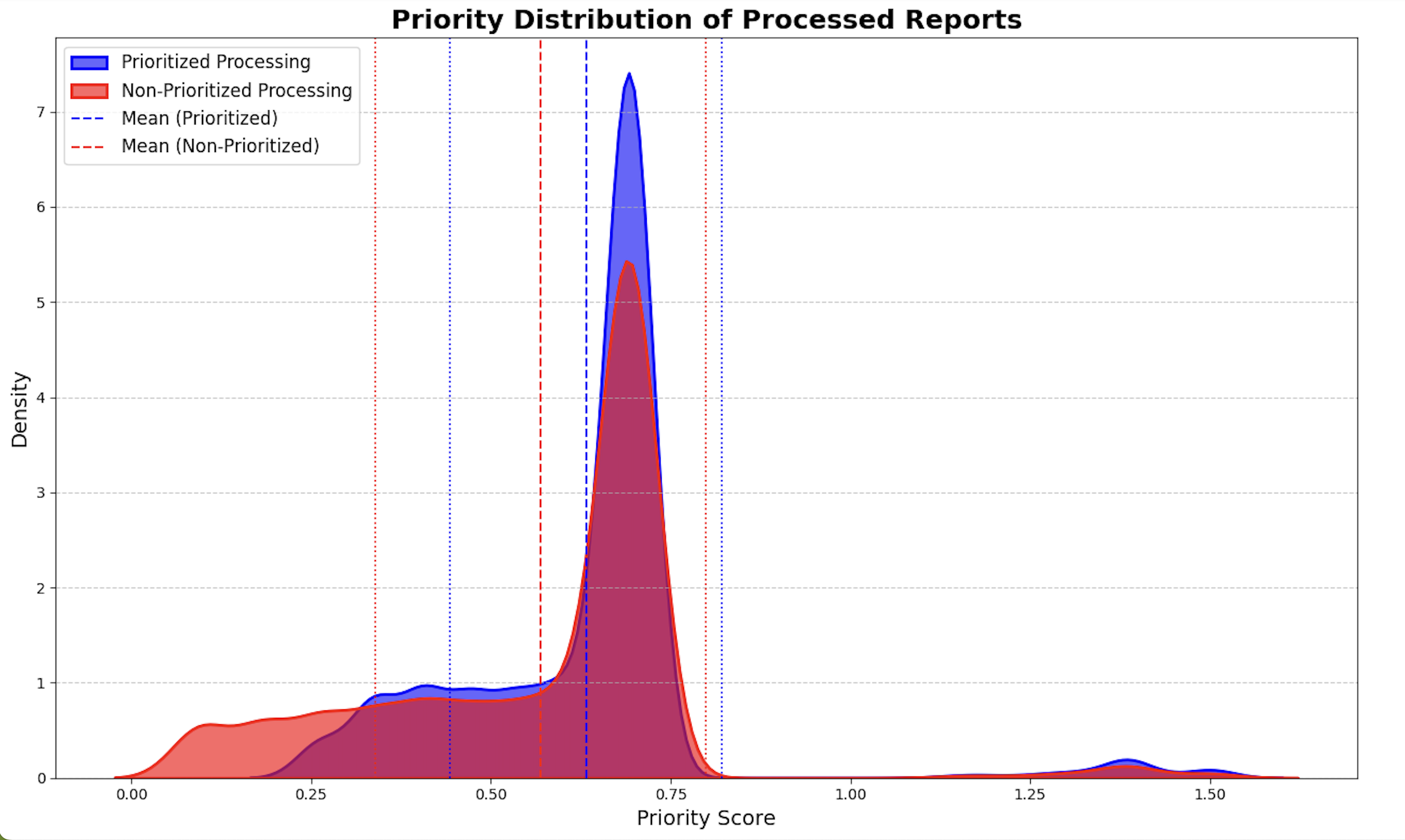}
    \caption{Priority Distribution of Processed Reports. The plot compares the density of priority scores between \textit{priority-based} and \textit{non-prioritised} processing. Dashed lines indicate the mean priority for each processing method, highlighting the effectiveness of prioritisation in focusing on critical cases.}
    \label{fig:priority_distribution}
\end{figure}

The ToxicChat dataset, derived from the Vicuna online demo, is a high-quality, crowdsourced benchmark for studying toxicity and safety in AI interactions \cite{lin2023toxicchat}. The developers aimed to obtain a realistic evaluation of their newly developed model by leveraging crowdsourced data, capturing real-world challenges in moderating harmful content submitted asynchronously and at scale. It contains 10,165 reports annotated for toxicity, jailbreaking, and OpenAI moderation scores, alongside detailed human annotations.

To emulate these conditions, we simulated Vicuna's internal prioritisation schema, where moderation operates under constraints such as supervision costs and limited capacity. Supervision costs were derived from the complexity of annotations, while accessibility was estimated from user interaction data. Reports were prioritised using a weighted combination of toxicity severity and accessibility, ensuring that high-priority cases were addressed first. Non-prioritised processing followed a first-come, first-served approach.

Figure~\ref{fig:priority_distribution} illustrates the priority score distributions of processed reports under both strategies. Importantly, the distributions only include reports that were successfully processed within the simulation period and do not include those that remained in the backlog. Reports processed via prioritisation had a narrower score distribution (mean: 0.63, SD: 0.19), reflecting a focus on critical cases with higher priority. In contrast, non-prioritised processing had a broader distribution (mean: 0.57, SD: 0.23), reflecting the absence of targeted focus.

This analysis demonstrates the efficacy of prioritisation in handling high-risk reports efficiently, ensuring timely intervention while optimizing resource use. By focusing on reports with higher priority scores, prioritisation strategies effectively reduce the likelihood of critical cases being left unattended. The ToxicChat dataset, with its detailed annotations and real-world applicability, underscores the advantages of targeted moderation frameworks in addressing the challenges of crowdsourced AI moderation.

\end{document}